%% file: main_arxiv.tex
\pdfoutput=1
\documentclass[10pt,twocolumn,letterpaper]{article}
\usepackage[pagenumbers]{cvpr}
\input{preamble}
\usepackage{bbding}
\definecolor{cvprblue}{rgb}{0.21,0.49,0.74}
\usepackage[pagebackref,breaklinks,colorlinks,allcolors=cvprblue]{hyperref}
\hypersetup{
  pdftitle={SRED: Sparse Regional Evidence Distillation Across Magnifications in Whole-Slide Imaging},
  pdfauthor={Zhiyuan Yang, Jiahao Cheng, Vincent Quoc-Huy Trinh, Mahdi S. Hosseini},
}

\makeatletter
\newif\ifSRED@suppaliases
\newcommand*{\SREDSuppLabelAliases}{\SRED@suppaliasestrue}
\AtEndPreamble{%
  \let\SRED@kernel@newlabel\newlabel
  \def\newlabel#1#2{%
    \SRED@kernel@newlabel{#1}{#2}%
    \ifSRED@suppaliases\SRED@kernel@newlabel{supp-#1}{#2}\fi}%
}
\AtBeginDocument{%
  \global\SRED@suppaliasesfalse
  \if@filesw
    \immediate\write\@auxout{\string\providecommand\string\SREDSuppLabelAliases{}}%
  \fi
}
\newcommand*{\SREDStartSupplement}{%
  \clearpage
  \if@filesw\immediate\write\@auxout{\string\SREDSuppLabelAliases}\fi
}
\makeatother

\def\confName{CVPR}
\def\confYear{2027}
\title{\methodname{}: Sparse Regional Evidence Distillation Across Magnifications in Whole-Slide Imaging}
\author{%
  Zhiyuan Yang\textsuperscript{1,3}\qquad
  Jiahao Cheng\textsuperscript{1,3}\qquad
  Vincent Quoc-Huy Trinh\textsuperscript{2}\qquad
  Mahdi S. Hosseini\Envelope\textsuperscript{1,3}\\[0.4em]
  \normalsize\textsuperscript{1}Department of Computer Science and Software Engineering (CSSE), Concordia University, Montreal, Canada\\
  \normalsize\textsuperscript{2}Axe Cancer, Centre de recherche du CHUM, Universit\'e de Montr\'eal, Montr\'eal, Canada\\
  \normalsize\textsuperscript{3}Mila - Quebec AI Institute, Montreal, Canada%
}
\begin{document}
\maketitle
\makeatletter
\begingroup
\renewcommand{\thefootnote}{\Envelope}%
\protected@xdef\@thefnmark{\thefootnote}%
\@ifundefined{H@@footnotetext}{\@footnotetext}{\H@@footnotetext}%
  {Corresponding author: \texttt{mahdi.hosseini@concordia.ca}}%
\endgroup
\makeatother
\input{main/0_abstract}

\begin{figure}[!t]
  \centering
  \includegraphics[width=\columnwidth]{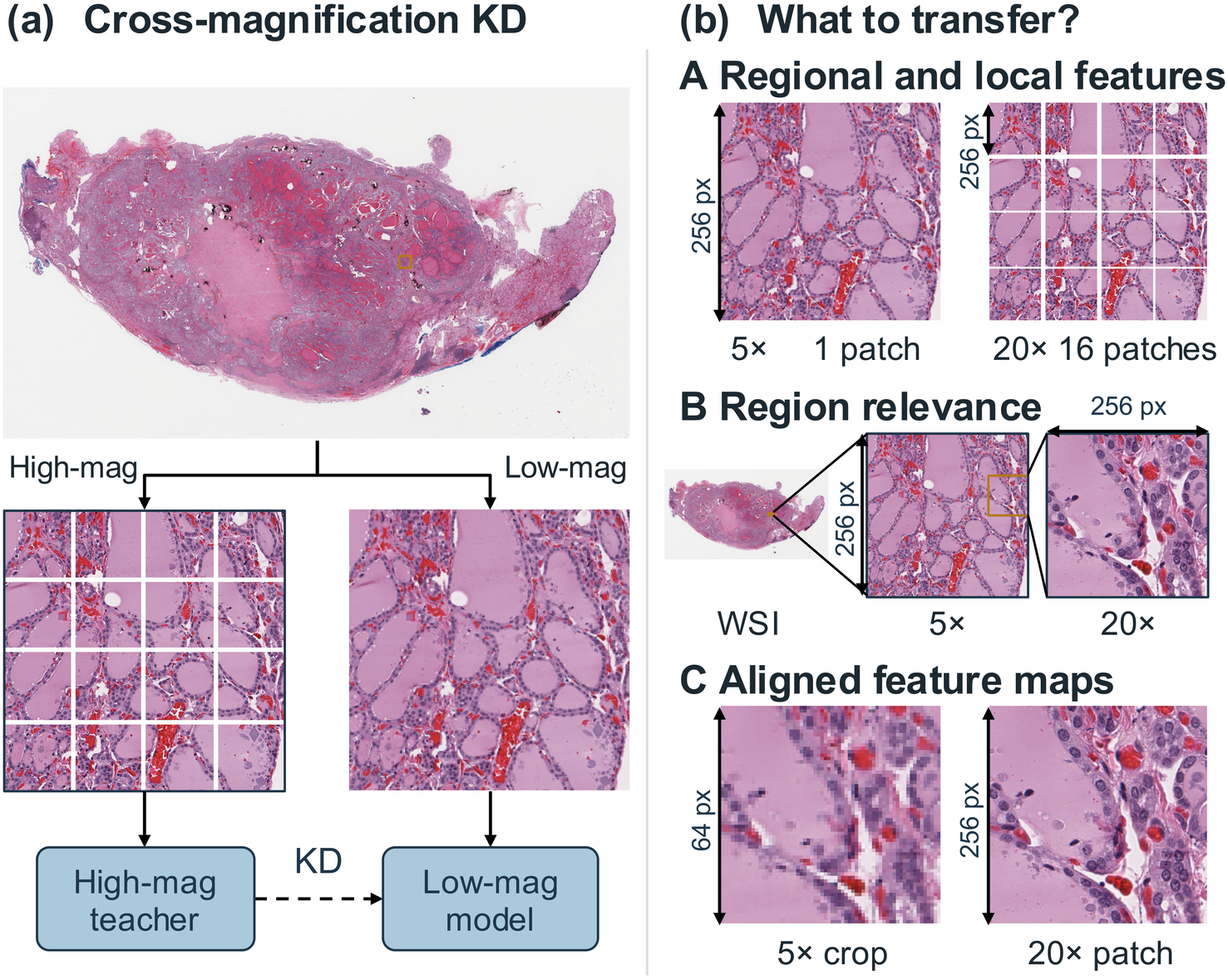}
  \caption{\textbf{WSI Cross-magnification knowledge transfer}. (a) High-magnification teacher guides a model learning from lower-magnification inputs. (b) Representative transfer strategies: (A) transfers regional and local features \cite{su2026xmag}; (B) transfers region relevance to a low-mag selector \cite{dong2025hdmil}; and (C) matches feature maps and pooled representations from different tissue resolutions \cite{han2026mag}.}
  \label{fig:cross_mag_background}
\end{figure}
\begin{figure*}[!t]
  \centering
  \includegraphics[width=\textwidth]{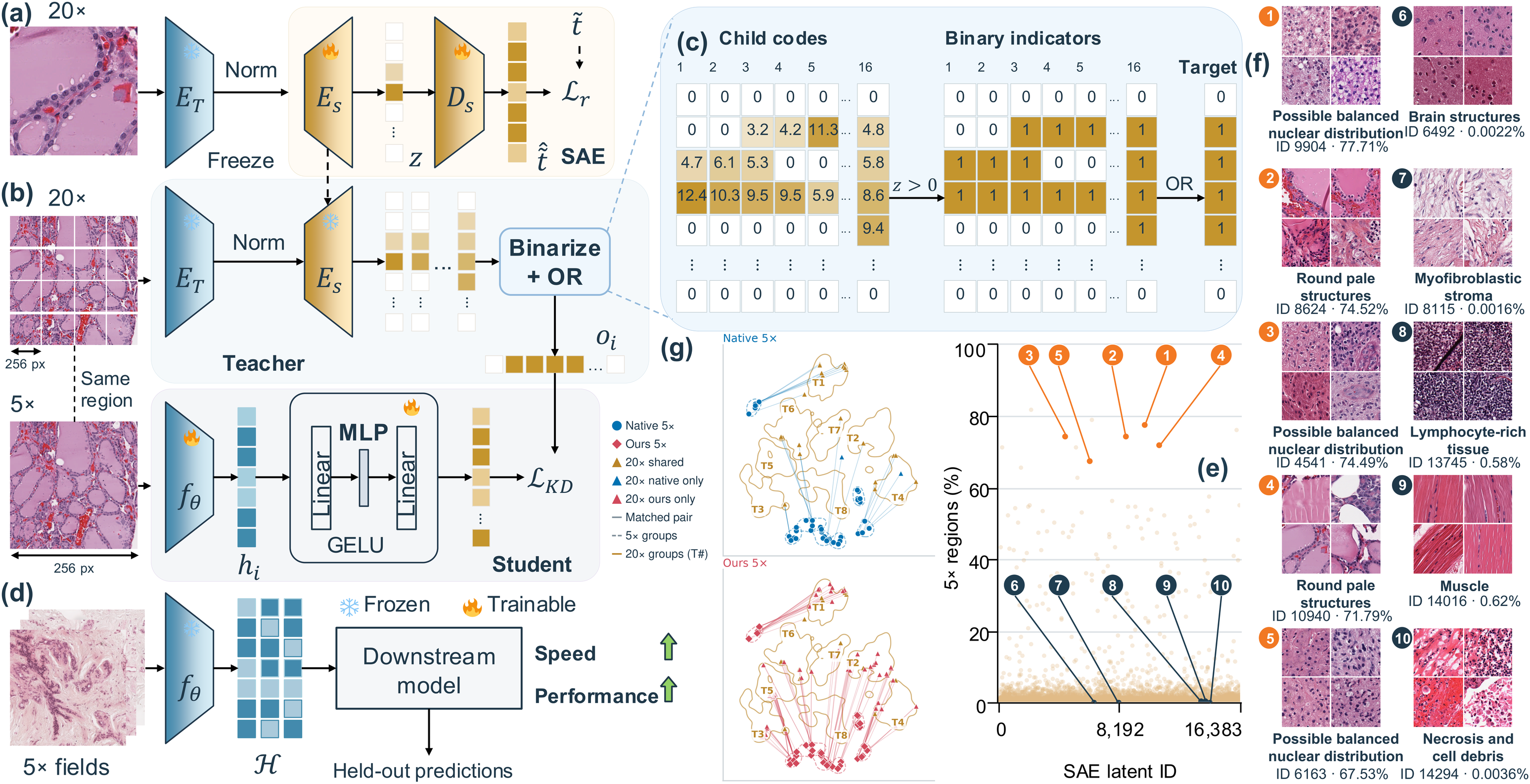}
  \caption{\textbf{\methodname{} transfers regional sparse evidence from \twenty{} children to a \five{} encoder.} \textbf{(a)} An SAE learns a sparse dictionary from \twenty{} UNI2-h features. \textbf{(b)} A \five{} encoder learns to predict regional events detected in \twenty{} patches. \textbf{(c)} Positive latent activations for each aligned \five{} region are binarized and combined by logical OR to form regional event targets. \textbf{(d)} The distilled encoder supplies \five{} patch features for downstream tasks. \textbf{(e)} The plot shows the percentage of \five{} regions in which each event occurs from SAE training. \textbf{(f)} Each group shows activating tissue examples with pathologist descriptions for latents labelled in (e). \textbf{(g)} The t-SNE compares native and distilled \five{} features and their links to aligned \twenty{} children on one TCGA-KIRC slide.}
  \label{fig:architecture}
\end{figure*}
\input{main/1_intro}

\input{main/2_related_work}

\input{main/3_method}
\input{main/5_experiments}
\input{main/6_results}

\input{main/7_ablations}

\input{main/10_conclusion}
\input{main/11_llm_usage}
{\small
\bibliographystyle{ieeenat_fullname}
\bibliography{main}
}

\SREDStartSupplement
\maketitlesupplementary
\appendix
\input{supplementary/A_implementation}
\input{supplementary/B_method_details}

\input{supplementary/C_datasets}

\FloatBarrier
\input{supplementary/D_additional_results}

\input{supplementary/E_qualitative}
\FloatBarrier
\end{document}

%% file: preamble.tex
\usepackage[T1]{fontenc}
\usepackage{graphicx}
\usepackage{amsmath,amssymb,amsthm}
\usepackage{booktabs}
\usepackage{colortbl}
\definecolor{resultBest}{HTML}{8BC3A1}
\definecolor{resultSecond}{HTML}{B2D8BF}
\definecolor{resultThird}{HTML}{D0E7D8}
\usepackage{multirow}
\usepackage{placeins}
\usepackage{stfloats}

\newcommand{\five}{$5\times$}
\newcommand{\twenty}{$20\times$}
\newcommand{\methodname}{SRED}

%% file: main/0_abstract.tex
\begin{abstract}
In computational pathology, high-magnification whole-slide image analysis preserves tissue detail but requires encoding many patches, whereas lower magnification reduces computation by covering a higher field of view per patch while sacrificing detail. Existing cross-magnification distillation matches features that often encode detail unreachable by low-magnification input, sometimes requiring slide-level labels that limit scalability. To address those issues, we introduce Sparse Regional Evidence Distillation (\methodname{}), which transfers regional event occurrences that encode reachable high-level interactions between details without slide-level labels. We use a sparse autoencoder (SAE) to learn an event dictionary from \twenty{} teacher features. Logical OR pooling records whether each event is detected in any spatially aligned patch, retaining both localized and shared evidence within the corresponding \five{} region. A pretrained \five{} encoder is fine-tuned to predict these regional targets without slide-level labels. Visual analysis identifies dictionary components responsive to recognizable histological patterns, supporting target interpretation. Compared with native \five{} features, \methodname{} improves average macro-F1 by $1.1$ points across seven classification and grading cohorts and average C-index by $1.7$ points across five survival cohorts. Runtime matches native \five{}, with an average $11.58\times$ speedup over \twenty{}. 
\end{abstract}

%% file: main/1_intro.tex
\section{Introduction}
\label{sec:introduction}
The standard whole-slide image (WSI) analysis workflow is inefficient due to its exhaustive patch extraction and embedding procedure for the downstream models, most commonly processed at high magnification, e.g. \twenty{} \cite{hosseini2024computational,dong2025hdmil, MF-MIL}. A WSI can lead to tens of thousands of patches \cite{shin2026lrmil} and a remedy to that is to extract and embed at low magnification to reduce the number of patches. However, it harms downstream task performance due to the loss of pixel information, known as pixel resolution \cite{su2026xmag,han2026mag}.

Cross-magnification distillation aims to bridge this gap by distilling the high-magnification teacher knowledge of an encoder into its low-magnification student feature embeddings, typically from \twenty{} to \five{}.  This builds a patch encoder that uses only low-magnification inputs to efficiently solve downstream tasks with high performance \cite{su2026xmag,shin2026lrmil,han2026mag} (see Fig.\ref{fig:cross_mag_background}). At equal patch sizes, one \five{} parent image patch covers a field of sixteen \twenty{} child patches, so distillation must relate a low-magnification representation to multiple high-magnification features. Existing approaches supervise the student through global or local teacher-feature alignment \cite{su2026xmag,han2026mag}, while some additionally transfer task-specific slide predictions and attention \cite{shin2026lrmil}. However, fine-grained teacher features may not be fully predictable from low-magnification inputs, and predicted features can under-represent variation in the teacher targets \cite{doesKDwork2021, yang2026transferable}.

To address those issues, we propose \textbf{Sparse Regional Evidence Distillation (\methodname{})}, which learns from visual events observed within each region instead of from aggregated high-magnification features. High-magnification features encode every fine-grained detail, while visual events tend to describe high-level relationships and interactions between the details that are both reachable and important for downstream tasks under low-magnification inputs. To detect the occurrence of observed high-magnification events, we fit a sparse autoencoder (SAE) on frozen high-magnification features \cite{gao2025scaling}. We treat a positive activation of a dictionary component,  which is usually monosemantic, as an event detection. A component enters the regional occurrence target if its code is positive in any observed child, including a response confined to a small part of the field. This provides proper exposure to rare events during training by assigning equal weights to all events. The target thus records component presence without averaging child feature vectors. During distillation, we fine-tune the student encoder to predict regional occurrence from low-magnification inputs, without needing to predict unreachable or unnecessary details. Our contributions are:
\begin{itemize}
    \item We introduce \methodname{}, a novel cross-magnification distillation framework that formulates knowledge transfer as prediction of regional event occurrence. It transfers evidence from aligned \twenty{} patches without slide labels and requires only \five{} inputs at inference.
    \item We construct regional targets from SAE activations using logical OR pooling, retaining both localized and shared event detections. The learned dictionary enables inspection of the tissue evidence used for supervision.
    \item We evaluate \methodname{} on 12 clinical cohorts covering classification, grading and survival prediction. It outperforms the existing distillation baselines in subtype classification and improves survival prediction over native \five{}. These gains are achieved at essentially native \five{} inference cost, with an $11.58\times$ speedup and $93.4\%$ fewer network FLOPs than native \twenty{} processing.
\end{itemize}

%% file: main/2_related_work.tex
\section{Related Work}
\label{sec:related_work}

\noindent\textbf{Knowledge distillation in pathology.}
Knowledge distillation (KD) transfers knowledge from teacher models to students by supervising their predictions or intermediate representations~\cite{hinton2015distilling,romero2015fitnets}. In computational pathology, KDTP~\cite{javed2023histology} uses multi-layer feature distillation to improve compact networks for tissue phenotyping, while resolution-based distillation~\cite{dipalma2021resolution} transfers knowledge from high-resolution teachers to lower-resolution students to reduce computational cost. Beyond efficiency, MoMA~\cite{levuong2025moma} employed contrastive distillation to adapt pretrained models to new pathology domains and tasks without accessing the original training data.
More recent approaches exploit complementary teachers: HistoMILKD~\cite{mallya2026histomilkd} combines knowledge from multiple pathology foundation models through a trainable MIL adapter for whole-slide classification. MKD~\cite{zhang2025multimodal} also uses paired genomic and histological data during training to improve biomarker prediction from histology alone at inference. 
These approaches transfer model predictions, feature representations or relationships between samples, without explicitly formulating supervision for information distributed across several patches. Our work addresses this target-design problem through regional event occurrence on high-magnification patches and OR-pooling aggregation.

\noindent\textbf{Cross-magnification knowledge transfer.}
In pathology, cross-magnification distillation also changes the tissue detail available to the student. MAG~\cite{han2026mag} aligns encoder features between magnifications before slide-level learning, and XMAG~\cite{su2026xmag} combines global and local teacher-feature supervision for a compact \five{} encoder. LRMIL~\cite{shin2026lrmil} aligns high-resolution child features with low-resolution spatial tokens, followed by distillation of slide predictions and attention. HDMIL~\cite{dong2025hdmil} instead distils high-resolution relevance masks into a low-resolution region selector and retains high-resolution classification on the selected tissue. Our work provides a new target for distillation, using regional event occurrence to transfer high-magnification evidence to a low-magnification patch encoder.

\noindent\textbf{Sparse autoencoders for visual representations.}
SAEs reconstruct dense model representations as sparse combinations of learned dictionary directions~\cite{gao2025scaling}. Each coordinate records a feature response that can be inspected and manipulated. In vision, CB-SAE~\cite{kulkarni2026cbsae} augments learned dictionaries with user-specified concepts to support interpretation and steering. ProtoMIL~\cite{sun2025protomil} applies SAE-derived concepts to pathology, expressing slide predictions as additive contributions that support concept-level interventions. Sparse Representation Matching~\cite{tran2021srm} uses teacher dictionary representations to construct local and image-level distillation targets. 
These studies establish sparse representations as tools for interpretation, prediction and distillation. However, spatially averaged sparse coefficients do not explicitly record the union of events across aligned high-magnification patches. We use OR-pooling on SAE codes to construct this regional occurrence target to train a low-magnification encoder for downstream tasks.

%% file: main/3_method.tex
\section{Method}
\label{sec:method}

Our SRED learns patch representations in two stages (Fig.~\ref{fig:architecture}). First, we capture the high-magnification evidence/events by training an SAE on \twenty{} teacher features to decompose them into a sparse dictionary.
For each \five{} patch, we aggregate event detections from the sparse codes of its aligned \twenty{} patches into a regional occurrence target. We transfer this high-magnification evidence by fine-tuning a pretrained UNI2-h encoder~\cite{chen2024uni} with a temporary head to predict each target from the corresponding \five{} patch. After distillation, we use the encoder to produce embeddings from \five{} patches for downstream evaluation.

\subsection{Decomposing Features into Events}
\label{sec:dictionary}

We train an SAE to learn a sparse dictionary from \twenty{} patch embeddings to decompose them into sparse semantics, which removes unnecessary information and keeps only high-level semantics as events (Fig.~\ref{fig:architecture}a). First, feature embeddings are extracted by the UNI2-h encoder~\cite{chen2024uni} that maps each patch $y$ to $t=E_T(y)\in\mathbb R^d$. These features are normalized as $\widetilde t=\mathcal N(t)$ using z-score normalization. The encoder maps $\widetilde t$ to a sparse code $z\in\mathbb R^{d_{\mathrm{SAE}}}$, and the decoder reconstructs the input from this code, such that
\begin{equation}
  z=\mathcal{S}_k\!\left(E(\widetilde t-b_{\mathrm{dec}})+c\right),
  \quad \widehat{\widetilde t}=Dz+b_{\mathrm{dec}},
  \label{eq:sae}
\end{equation}
where $E\in\mathbb R^{d_{\mathrm{SAE}}\times d}$ and $D\in\mathbb R^{d\times d_{\mathrm{SAE}}}$ are the encoder and decoder matrices, $c\in\mathbb R^{d_{\mathrm{SAE}}}$ is the encoder bias, and $b_{\mathrm{dec}}\in\mathbb R^d$ is a learned decoder bias, which we also subtract before encoding. Each coordinate $z_\ell$ is a latent activation that weights the dictionary vector in column $\ell$ of $D$. A positive activation, $z_\ell>0$, denotes the occurrence of the event associated with latent $\ell$. Each dictionary vector has Euclidean length one, so its activation controls the magnitude of its contribution to the reconstruction. The TopK activation $\mathcal S_k$~\cite{gao2025scaling} selects the $k$ largest encoder scores for each input, applies ReLU to these scores, and sets all other entries to zero, giving at most $k$ positive activations. To increase reconstruction capacity while keeping each code sparse, we use $d_{\mathrm{SAE}}=16{,}384$ dictionary vectors and $k=32$. We also adopt the auxiliary reconstruction loss (AuxK)~\cite{gao2025scaling}, which trains inactive latents to reconstruct the residual error during SAE training. The SAE objective is
\begin{equation}
  \begin{aligned}
    \mathcal{L}_{\mathrm{SAE}}
    ={}&\mathbb{E}\!\left[\frac{1}{d}
        \|\widetilde t-\widehat{\widetilde t}\|_2^2\right] +\lambda_{\mathrm{aux}}
      \underbrace{\mathbb{E}\!\left[\frac{1}{d}
        \|r-Dz_{\mathrm{aux}}\|_2^2\right]}_{\mathcal L_{\mathrm{AuxK}}},
  \end{aligned}
  \label{eq:sae-reconstruction}
\end{equation}
where $z_{\mathrm{aux}}$ is formed by applying TopK with a separate limit $k_{\mathrm{aux}}$ to the encoder scores of inactive latents, $r=\operatorname{sg}(\widetilde t-\widehat{\widetilde t})$ is the reconstruction residual, $\operatorname{sg}$ denotes stop-gradient, and $\lambda_{\mathrm{aux}}$ is the auxiliary loss coefficient.

\subsection{Constructing Regional Teacher Target}
\label{sec:regional-distillation}

Since each \five{} patch corresponds to a region of up to sixteen \twenty{} patches, we aggregate the per-patch event occurrence codes from \twenty{} into a single teacher target for the \five{} patch using \textbf{OR pooling}, which binarizes the occurrence codes and applies the logical OR operation (Fig.~\ref{fig:architecture}c). Let $x_i$ denote the \five{} patch and $\{y_{ij}\}_{j=1}^{n_i}$ its \twenty{} children from a $4\times4$ grid, where $1\leq n_i\leq16$ after tissue filtering. With the teacher encoder and SAE frozen, Eq.~\eqref{eq:sae} produces a code $z_{ij}$ for each child's teacher feature.
OR pooling applies the logical OR operator to the binary indicators of each event across the aligned \twenty{} patches. The resulting binary regional occurrence target $o_i\in\{0,1\}^{d_{\mathrm{SAE}}}$ is given by
\begin{equation}
  o_{i\ell}=\max_{1\leq j\leq n_i}\mathbf{1}\{z_{ij\ell}>0\},
  \qquad \ell=1,\ldots,d_{\mathrm{SAE}},
  \label{eq:occurrence-target}
\end{equation}
where $\mathbf{1}\{\cdot\}$ is the indicator function.
Binarization discards the occurrence strengths of events, and the OR operation discards the frequency of event occurrence within the \five{} region. This effectively assigns equal weights to all regional events and keeps rare events as exposed as common ones.

\subsection{Cross-magnification Knowledge Transfer}

We use the same pretrained teacher encoder as the student $f_\theta$ and fine-tune it on \five{} patches using the aggregated regional occurrence targets, which distills the encoder's high-magnification knowledge into its own low-magnification representations. We attach an MLP head to the encoder, which predicts $o_i$ from the student's patch embedding $h_i=f_\theta(x_i)\in\mathbb R^{d_s}$ and is discarded after training. Keeping this mapping in a separate head allows downstream models to use the backbone features without being restricted to the dictionary's occurrence probabilities. This MLP has two linear layers separated by a GELU activation~\cite{hendrycks2016gelu}.
We minimize binary cross-entropy (BCE) between the predicted occurrence probabilities $q_i=q_\phi(h_i)$ and the regional targets $o_i$, averaged over the $d_{\mathrm{SAE}}$ latent coordinates, which gives
\begin{equation}
  \begin{aligned}
    \mathcal{L}_{\mathrm{occ}}
    =-\mathbb{E}_{i}\frac{1}{d_{\mathrm{SAE}}}\sum_{\ell=1}^{d_{\mathrm{SAE}}}
    \big[&o_{i\ell}\log q_{i\ell}+(1-o_{i\ell})\log(1-q_{i\ell})\big].
  \end{aligned}
  \label{eq:occurrence-kd}
\end{equation}
The occurrence loss supervises predictions through the MLP, while downstream models use $h_i$ directly. To limit catastrophic forgetting of native \five{} knowledge, we adopt the cosine feature anchor loss~\cite{hou2019learning}. The anchor penalizes the cosine dissimilarity between $h_i$ and $h_i^0$,
\begin{equation}
  \mathcal L_F=\mathbb E_i\!\left[
    1-cos(h_i, h_i^0)\right]
  \label{eq:feature-anchor}
\end{equation}
which constrains the direction of the backbone features without limiting their magnitudes. To adapt the MLP to the regional occurrence targets before fine-tuning the encoder, we begin with an MLP warm-up. The student encoder remains frozen during this phase, and only the MLP is optimized with $\mathcal L_{\mathrm{occ}}$. After warm-up, we unfreeze the encoder and introduce the feature anchor for joint optimization of the encoder and MLP, which gives
\begin{equation}
  \mathcal L_{\mathrm{KD}}=\mathcal L_{\mathrm{occ}}+\lambda_F\mathcal L_F,
  \label{eq:total-kd}
\end{equation}
where $\lambda_F$ is a hyperparameter for the feature anchoring strength. During joint training, $\mathcal L_{\mathrm{occ}}$ updates both the encoder and MLP, while $\mathcal L_F$ acts only on the encoder (see Supplementary~\ref{supp-sec:supp_method}).

\subsection{Downstream Evaluation}
\label{sec:deployment}
SAE training and distillation produce a fine-tuned encoder that takes only \five{} patches as input. We evaluate its representations by freezing the encoder and training downstream models with slide or case labels (Fig.~\ref{fig:architecture}d). For a slide containing $N$ patches, the extracted embeddings form
\begin{equation}
  \mathcal H=[h_1;\ldots;h_N]\in\mathbb{R}^{N\times d_s}.
  \label{eq:slide-bag}
\end{equation}
These embeddings can be used as inputs to different downstream models. In the following experiments, we choose ABMIL~\cite{ilse2018attention} and linear probing for evaluation. ABMIL learns weights to aggregate the variable number of patch embeddings into a slide representation for classification, grading or survival. For linear probing, we average the patch embeddings and fit logistic regression to assess the information accessible through a linear predictor.

%% file: main/5_experiments.tex
\begin{table*}[!t]
\centering
\footnotesize
\begin{tabular*}{\textwidth}{@{\extracolsep{\fill}}llccccc}
\toprule
& & \multicolumn{2}{c}{Native} & \multicolumn{3}{c}{Cross-magnification KD methods} \\
\cmidrule(lr){3-4}\cmidrule(lr){5-7}
Dataset & Metric & \five{} & \twenty{} & XMAG~\cite{su2026xmag} & LRMIL~\cite{shin2026lrmil} & \methodname{} (ours) \\
\midrule
CAM16 & AUROC & \cellcolor{resultSecond}$0.9773_{\pm0.0146}$ & \cellcolor{resultThird}$0.9719_{\pm0.0247}$ & $0.9275_{\pm0.0195}$ & $0.8897_{\pm0.0378}$ & \cellcolor{resultBest}$\underline{\mathbf{0.9905}}_{\pm0.0105}$ \\
CAM17 & Macro-F1 & \cellcolor{resultThird}$0.5894_{\pm0.0310}$ & \cellcolor{resultBest}$\mathbf{0.6705}_{\pm0.0375}$ & $0.5508_{\pm0.0535}$ & $0.5118_{\pm0.0318}$ & \cellcolor{resultSecond}$\underline{0.6116}_{\pm0.0515}$ \\
Private-CRC & Macro-F1 & $0.8105_{\pm0.0124}$ & \cellcolor{resultSecond}$0.8229_{\pm0.0187}$ & \cellcolor{resultThird}$0.8191_{\pm0.0204}$ & $0.7794_{\pm0.0356}$ & \cellcolor{resultBest}$\underline{\mathbf{0.8249}}_{\pm0.0188}$ \\
BRACS & Macro-F1 & \cellcolor{resultBest}$\mathbf{0.4375}_{\pm0.0399}$ & \cellcolor{resultThird}$0.3875_{\pm0.0533}$ & $0.3825_{\pm0.0209}$ & $0.3021_{\pm0.1228}$ & \cellcolor{resultSecond}$\underline{0.4273}_{\pm0.0246}$ \\
PANDA & QW $\kappa$ & \cellcolor{resultThird}$0.8367_{\pm0.0078}$ & \cellcolor{resultBest}$\mathbf{0.8661}_{\pm0.0046}$ & $0.8247_{\pm0.0018}$ & $0.8181_{\pm0.0037}$ & \cellcolor{resultSecond}$\underline{0.8587}_{\pm0.0062}$ \\
\midrule
IMP & Macro-F1 & \cellcolor{resultThird}$0.8991_{\pm0.0019}$ & \cellcolor{resultSecond}$0.9031_{\pm0.0050}$ & $0.8937_{\pm0.0046}$ & $0.8926_{\pm0.0024}$ & \cellcolor{resultBest}$\underline{\mathbf{0.9056}}_{\pm0.0028}$ \\
IMP-Cervix & Macro-F1 & \cellcolor{resultBest}$\mathbf{0.7096}_{\pm0.0337}$ & \cellcolor{resultThird}$0.6471_{\pm0.0630}$ & $0.6076_{\pm0.0660}$ & $0.5607_{\pm0.0408}$ & \cellcolor{resultSecond}$\underline{0.6884}_{\pm0.0213}$ \\
\midrule
TCGA-KIRC & C-index & \cellcolor{resultSecond}$0.7158_{\pm0.0292}$ & \cellcolor{resultThird}$0.7154_{\pm0.0380}$ & \textcolor{red}{$0.6969_{\pm0.0250}$} & $0.6797_{\pm0.0551}$ & \cellcolor{resultBest}$\underline{\mathbf{0.7258}}_{\pm0.0301}$ \\
TCGA-KIRP & C-index & \cellcolor{resultThird}$0.6184_{\pm0.1486}$ & \cellcolor{resultBest}$\mathbf{0.6485}_{\pm0.0903}$ & \textcolor{red}{$\underline{0.6471}_{\pm0.0530}$} & $0.6154_{\pm0.0451}$ & \cellcolor{resultSecond}$0.6294_{\pm0.0985}$ \\
TCGA-LUAD & C-index & $0.5165_{\pm0.0283}$ & \cellcolor{resultBest}$\mathbf{0.5796}_{\pm0.0342}$ & \textcolor{red}{$\underline{0.5791}_{\pm0.0157}$} & \cellcolor{resultSecond}$0.5739_{\pm0.0669}$ & \cellcolor{resultThird}$0.5611_{\pm0.0336}$ \\
TCGA-STAD & C-index & $0.5288_{\pm0.0501}$ & \cellcolor{resultBest}$\mathbf{0.5600}_{\pm0.0387}$ & \textcolor{red}{$\underline{0.5455}_{\pm0.0106}$} & \cellcolor{resultSecond}$0.5435_{\pm0.0580}$ & \cellcolor{resultThird}$0.5376_{\pm0.0599}$ \\
TCGA-UCEC & C-index & \cellcolor{resultSecond}$0.7950_{\pm0.0473}$ & \cellcolor{resultThird}$0.7525_{\pm0.0926}$ & \textcolor{red}{$0.7802_{\pm0.0556}$} & $0.7406_{\pm0.0833}$ & \cellcolor{resultBest}$\underline{\mathbf{0.8040}}_{\pm0.0288}$ \\
\bottomrule
\end{tabular*}
\caption{\textbf{Clinical benchmark comparison.} Entries are $\mathrm{mean}_{\pm\mathrm{SD}}$, with sample SD over five seeds. Native \five{} and Native \twenty{} use UNI2-h~\cite{chen2024uni} features at the corresponding magnification, and XMAG uses its released encoder. LRMIL is our local reproduction of its paper setting. Higher values are better for every metric. Bold marks the best mean across all methods, and underlining marks the best among XMAG, LRMIL and \methodname{}. For XMAG~\cite{su2026xmag}, red entries mark cohorts used for their encoder pretraining. Our SAE training and distillation exclude these data. Green backgrounds mark the top three means in each row, from darkest (best) to lightest (third), excluding XMAG for survival. Our method often matches the highest native performance closely, sometimes exceeding them.}
\label{tab:main_results}
\end{table*}

\section{Experimental design}
\label{sec:experiments}

We compare \methodname{} with native encoders and other cross-magnification distillation methods on 12 datasets covering classification, grading and survival prediction.
 Figures~\ref{fig:final_stage_distribution} and~\ref{fig:downstream_cohorts} summarize the datasets used for different stages.

\subsection{Dataset}
\label{sec:datasets_splits}

\begin{figure}[t]
\centering
\includegraphics[width=\linewidth]{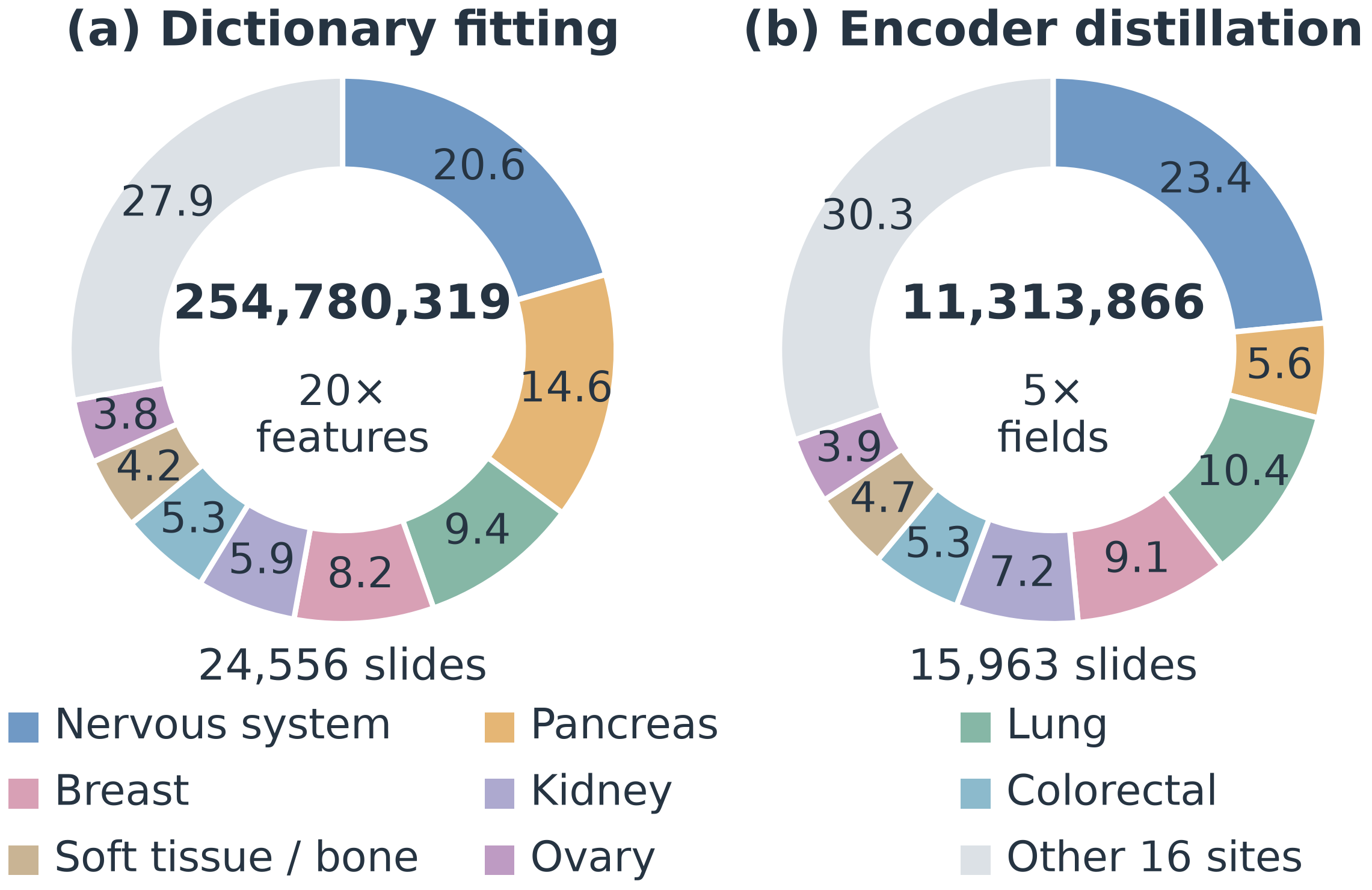}
\caption{\textbf{Dictionary fitting and distillation share anatomical coverage but use different inputs.} (a) SAE fitting uses 254,780,319 \twenty{} features from 24,556 slides. (b) Distillation uses 11,313,866 \five{} fields from 15,963 slides.}
\label{fig:final_stage_distribution}
\end{figure}

We train the SAE and fine-tune the student on TCGA~\cite{tcga_resource}, CPTAC~\cite{li2023cptac}, EBRAINS~\cite{roetzer2022ebrains}, MUT-HET-RCC~\cite{acosta2022heterogeneity}, OV-Bevacizumab~\cite{wang2022ovarian}, BC-therapy~\cite{sammut2022therapy}, NADT-Prostate~\cite{wilkinson2020nadt}, the H\&E/MSKCC subset of BOEHMK~\cite{boehm2022multimodal}, and MBC~\cite{bergstrom2024hrd,galland2022platinum}, together with additional in-house data. These collections span 24 anatomical groups (Fig.~\ref{fig:final_stage_distribution}). We use AtlasPatch~\cite{atlaspatch} for patch extraction and feature embedding.

\begin{figure}[t]
\centering
\includegraphics[width=\linewidth]{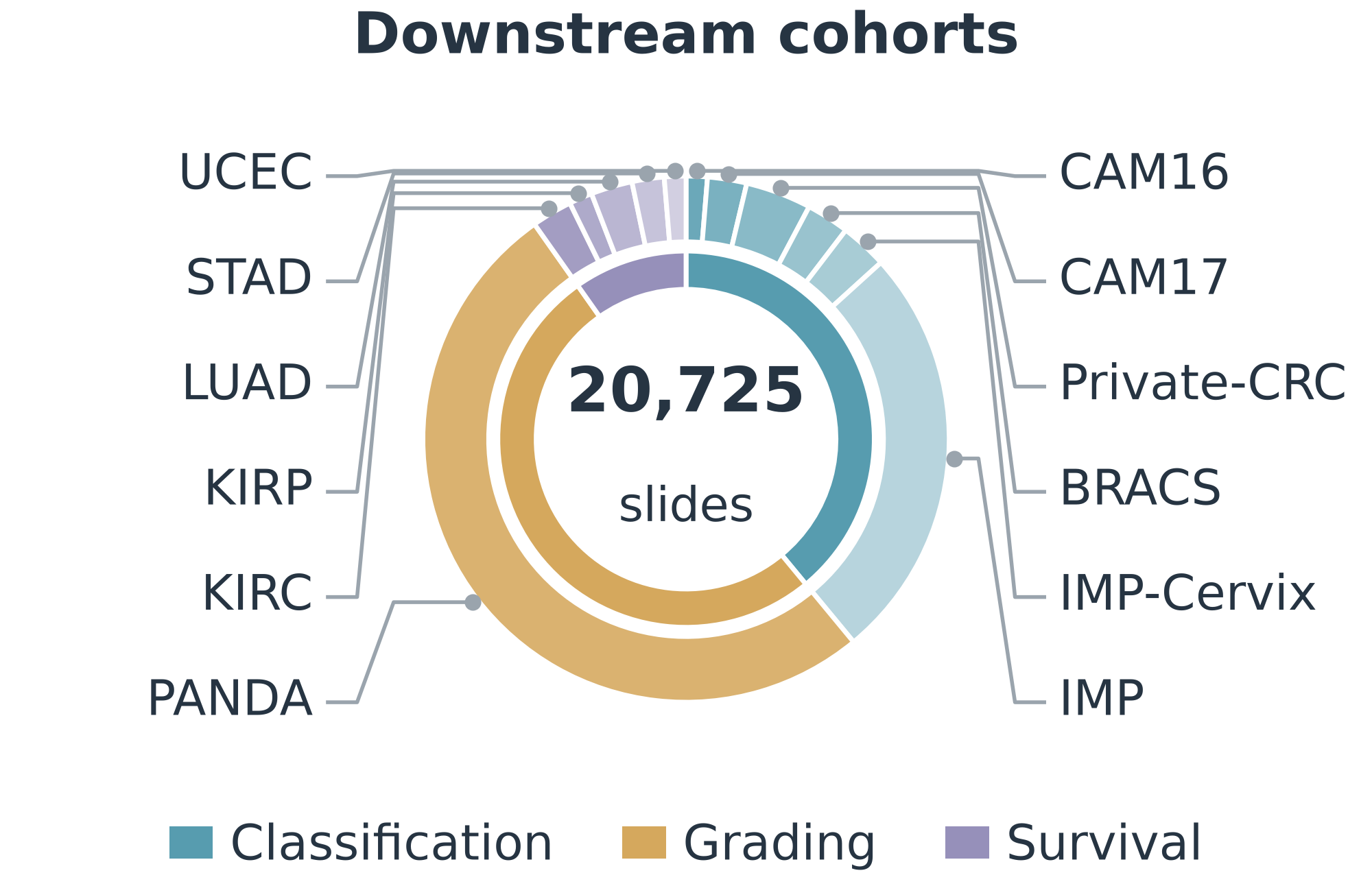}
\caption{\textbf{Downstream cohorts cover classification, grading and survival.}
The inner ring groups tasks; the outer ring groups twelve cohorts, with areas proportional to their 20,725 labeled slides.
 KIRC/KIRP/LUAD/STAD/UCEC denote TCGA cohorts.}
\label{fig:downstream_cohorts}
\end{figure}

Downstream evaluation uses completely held-out datasets, including CAM16 and CAM17~\cite{litjens2018camelyon}, Private-CRC, BRACS~\cite{brancati2022bracs}, PANDA~\cite{bulten2022panda}, IMP~\cite{neto2024impcrs}, IMP-Cervix~\cite{oliveira2024impcervix}, and five TCGA survival cohorts: KIRC, KIRP, LUAD, STAD and UCEC~\cite{tcga_resource,liu2018clinical} (Fig.~\ref{fig:downstream_cohorts}).
 These datasets are excluded from both SAE training and student encoder fine-tuning. See Supplementary Sec.~\ref{supp-sec:supp_data} for more details.

\subsection{Baselines and downstream models}
\label{sec:comparison_models}
We compare against the encoder's own native performance and against other KD methods. The native baselines use the frozen pretrained UNI2-h~\cite{chen2024uni} at \five{} and \twenty{}.  For XMAG~\cite{su2026xmag}, we use their released checkpoint and follow their exact embedding extraction protocol at \five{}. For LRMIL~\cite{shin2026lrmil}, we follow the paper setting that has two-stage distillation with slide-level label information, where the first stage transfers patch token features from CLS tokens and the aligned spatial tokens, and the second stage performs slide-level distillation using slide-level labels. For survival analysis, LRMIL formulates survival prediction as binary classification of vital status~\cite{shin2026lrmil}. To account for follow-up time and censoring in our survival tasks, we adapt its MIL models to the discrete-time survival formulation~\cite{chen2022pancancer}.
On IMP~\cite{neto2024impcrs} and IMP-Cervix~\cite{oliveira2024impcervix}, all methods use mean pooling over feature bags followed by logistic regression. LRMIL uses only its patch-level distillation stage. For the remaining tasks, we use ABMIL~\cite{ilse2018attention} as the downstream model. See Supplementary Sec.~\ref{supp-sec:supp_implementation} for more details.

\subsection{Evaluation metrics}
\label{sec:evaluation_metrics}
We report AUROC for CAM16~\cite{bejnordi2017camelyon16} and macro-F1 for the other classification tasks. For PANDA~\cite{bulten2022panda}, we use quadratic weighted kappa, while survival prediction is evaluated using Harrell's C-index~\cite{harrell1982evaluating}. For cases with multiple slides, we average the predicted slide risks before calculating the C-index. We report the mean and sample standard deviation (SD) over five seeds on the evaluation sets.

%% file: main/6_results.tex
\section{Results}
\label{sec:results}

\subsection{Clinical benchmark comparison}
\label{sec:clinical_comparison}
Table~\ref{tab:main_results} summarizes the clinical benchmark results across classification, grading and survival prediction. Our \methodname{} generally outperforms native \five{} features and the other distillation baselines, particularly on classification tasks.
It improves the average macro-F1 by $1.1$ points relative to native \five{} and by $0.4$ points relative to native \twenty{}.
 It achieves the best results on CAM16 and Private-CRC and leads the linear-probe comparison on IMP. On PANDA, it closes much of the gap between native \five{} and \twenty{}. Compared with others, \methodname{} also shows lower variation. On BRACS and IMP-Cervix, where native \five{} already outperforms native \twenty{}, our \methodname{} retains this advantage over \twenty{} and also outperforms the other distillation baselines.

For survival prediction, our \methodname{} improves on native \five{} across all cohorts, increasing the average C-index by $1.7$ points, with the largest gain on TCGA-LUAD. XMAG's encoder pretraining includes these TCGA cohorts~\cite{su2026xmag}, which may contribute to its strong performance, whereas \methodname{} are excluded them from training. Despite this difference in training exposure, \methodname{} still achieves the highest C-indices on TCGA-KIRC and UCEC. The gains on CAM16 and TCGA-UCEC further show that distillation using SAE occurrence targets can improve the \five{} representation even when native \five{} already outperforms native \twenty{}. Overall, using only \five{} inputs, our \methodname{} achieves close performance to native \twenty{}, with an $11.58\times$ speedup over native \twenty{} processing (see Supplementary Sec.~\ref{supp-sec:supp_results} for details).

\begin{figure*}[!t]
\centering
\includegraphics[width=\textwidth]{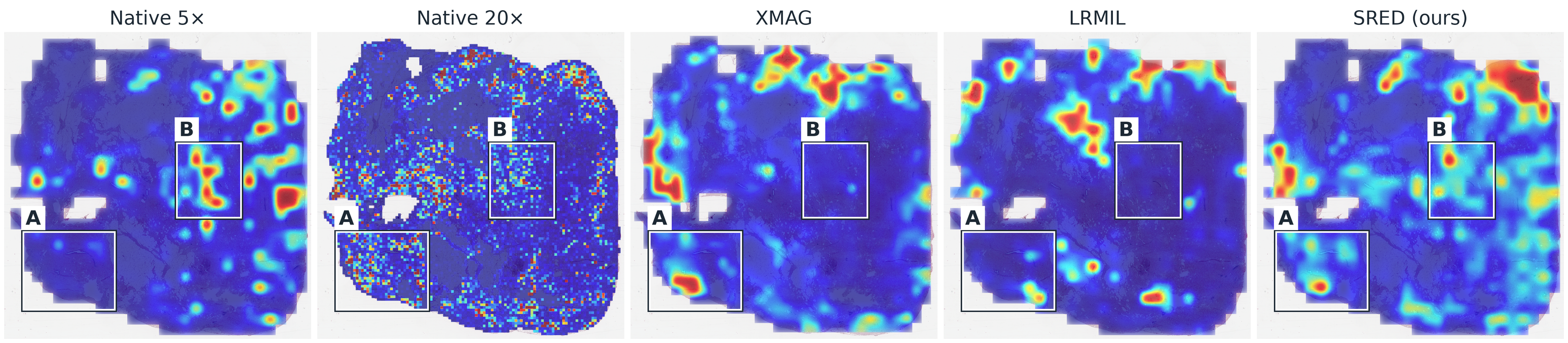}\\[2pt]
\includegraphics[width=\textwidth]{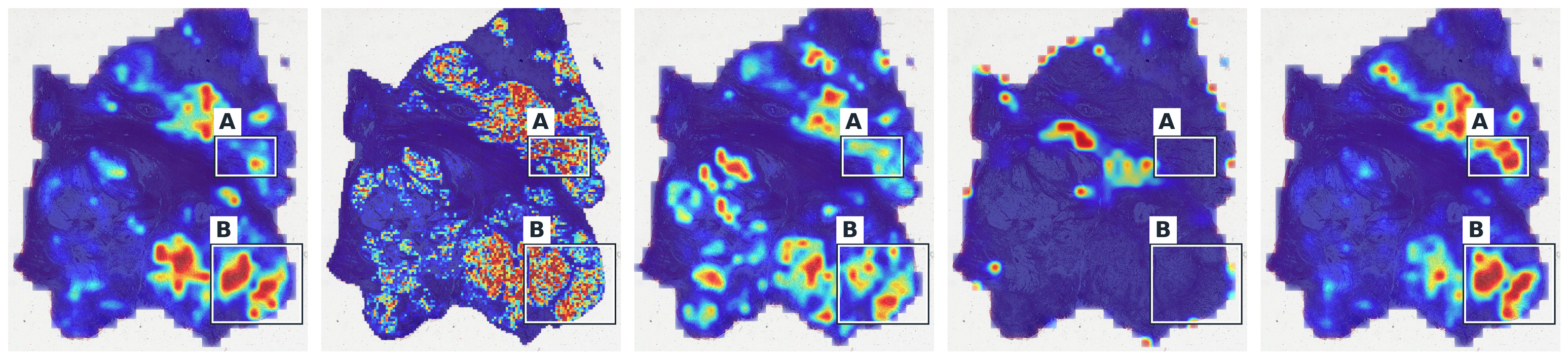}
\caption{\textbf{Attention heatmaps for classification and survival prediction.} BRACS (top) and TCGA-STAD (bottom) examples. Within each slide, regions A and B mark the same tissue locations across methods. Display scales are normalized separately for each method.}
\label{fig:bracs_attention}
\end{figure*}

\subsection{Qualitative analysis}
\label{sec:qualitative_analysis}
A senior pathologist assessed whether patches that activate the same SAE latent share the same histological patterns.
Fig.~\ref{fig:architecture}e shows that 98.9\% of events are detected in at most 10\% of \five{} training regions after OR pooling, while the most frequent event occurs in 81.9\%. Fig.~\ref{fig:architecture}f show four examples of activating patches for each latent. Among the less frequent examples, the pathologist identified specific events like myofibroblastic stroma, and lymphocyte-rich tissue. For the frequent examples, most are general visual patterns such as nuclear arrangements and pale rounded structures on an eosinophilic background. Other frequent latent examples had no clear histological pattern.

We also compare feature correspondence before and after distillation using the same \five{} regions.
Fig.~\ref{fig:architecture}g shows that on the TCGA-KIRC example, our \methodname{} retrieves aligned \twenty{} children from all eight teacher feature groups, where native \five{} has no matches in T3 or T5. The additional matches indicate broader correspondence with the teacher feature distribution for these regions.

Fig.~\ref{fig:bracs_attention} compares attention heatmaps for BRACS classification and TCGA-STAD survival prediction. In the BRACS example, compared with native \five{}, our \methodname{} shows higher attention in A and a less prominent but still distinct response in B. The closer correspondence with native \twenty{} in both regions suggests that the student learns regional cues from high-magnification supervision. XMAG highlights A but shows little response in B, whereas LRMIL does not clearly reproduce the native \twenty{} pattern in either region. On TCGA-STAD, our \methodname{} strengthens the response in region A, which is more prominent in native \twenty{}, while preserving the native \five{} focus in region B.

%% file: main/7_ablations.tex
\subsection{Ablation studies on SAE}
\label{sec:sae_ablations}
We ablate how dictionary width $d_{\mathrm{SAE}}$ and sparsity $k$ affect the SAE codes used to construct regional targets. Both comparisons use data containing approximately $5.4$ million \twenty{} features reserved for configuration comparison. We report mean squared reconstruction error (MSE) per feature coordinate and fraction of variance unexplained (FVU)~\cite{rajamanoharan2024jumprelu}. The FVU divides the total squared reconstruction error by the sum of squared deviations of the normalized features.
\begin{table}[!htbp]
\centering
\footnotesize
\setlength{\tabcolsep}{2pt}
\begin{tabular*}{\columnwidth}{@{\extracolsep{\fill}}lcccc@{}}
\toprule
Width $d_{\mathrm{SAE}}$
& MSE $\downarrow$
& FVU $\downarrow$
& \shortstack{Observed\\(\%)}
& \shortstack{Unobserved\\(\%)} \\
\midrule
$4{,}096$
& $0.2500$
& $0.2422$
& $100.0000$
& $0.0000$ \\
$8{,}192$
& $0.2360$
& $0.2286$
& $99.9756$
& $0.0244$ \\
\rowcolor{resultSecond}
$16{,}384$
& $0.2297$
& $0.2225$
& $96.9727$
& $3.0273$ \\
$32{,}768$
& $0.2273$
& $0.2202$
& $84.1431$
& $15.8569$ \\
\bottomrule
\end{tabular*}
\caption{\textbf{Dictionary width ablation.} Observed and unobserved report the percentages of dictionary components with and without a positive activation anywhere in this set. Green marks our selected setting.}
\label{tab:sae_width}
\end{table}

\noindent\textbf{Dictionary width.}
We vary $d_{\mathrm{SAE}}$ at $k=32$ to expand the available dictionary. Table~\ref{tab:sae_width} shows that wider dictionaries improve reconstruction. At $d_{\mathrm{SAE}}=16{,}384$, 96.97\% of components activate on the selection features. This provides substantially more active components than the smaller settings while leaving few components unobserved. Doubling the width to $32{,}768$ yields only a further 1.0\% relative reduction in MSE with 84\% active components. Thus, $d_{\mathrm{SAE}}=16{,}384$ provides a large pool of active components for regional supervision, with reconstruction close to the largest setting at half the dictionary and target width.

\noindent\textbf{Sparsity.}
We ablate $k$ at $d_{\mathrm{SAE}}=16{,}384$ to examine the balance between reconstruction and the strength of the activations used as event detections. This balance matters because OR pooling treats every activation as a detection regardless of its magnitude. Table~\ref{tab:sae_sparsity} shows that $k=32$ improves reconstruction over $k=16$ while retaining 76.37\% of positive coefficients above two. Increasing $k$ improves reconstruction, but shifts a substantially larger proportion of positive activations toward smaller coefficients. Thus, using $k=32$ balances reconstruction fidelity with a compact event code dominated by strong activations.

\begin{table}[!htbp]
\centering
\footnotesize
\setlength{\tabcolsep}{2pt}
\begin{tabular*}{\columnwidth}{@{\extracolsep{\fill}}lccccc@{}}
\toprule
& & \multicolumn{4}{c}{Positive activations (\%)} \\
\cmidrule(lr){3-6}
\shortstack{Active\\limit $k$}
& FVU $\downarrow$
& $0<z\leq0.5$
& $0.5<z\leq1$
& $1<z\leq2$
& $z>2$ \\
\midrule
$16$
& $0.2750$
& $0.0000$
& $0.0035$
& $1.3536$
& $98.6429$ \\
\rowcolor{resultSecond}
$32$
& $0.2225$
& $0.0001$
& $0.0653$
& $23.5641$
& $76.3705$ \\
$64$
& $0.1727$
& $0.0017$
& $2.9744$
& $52.2741$
& $44.7497$ \\
$128$
& $0.1279$
& $0.0707$
& $27.8718$
& $46.1693$
& $25.8881$ \\
\bottomrule
\end{tabular*}
\caption{\textbf{Sparsity ablation.} FVU and activation percentages are computed in the same set as Table~\ref{tab:sae_width}. Percentages use positive activations as the denominator. The intervals describe activation magnitudes and event detection remains defined by $z>0$.}
\label{tab:sae_sparsity}
\end{table}

\subsection{Ablation studies on KD}
\label{sec:kd_ablations}

For the distillation ablations, we train each student variant, including our method, on a shared subset of approximately $0.75$ million \five{} regions from $998$ slides that covers all 24 anatomical groups represented in the full distillation training. The regional-aggregation, anchoring and prediction-loss comparisons each vary one design choice while keeping the remaining configuration and downstream validation partitions fixed. We use ABMIL for classification on CAM16 and survival prediction on TCGA-STAD, and a linear probe on IMP. Additional KD ablations are reported in Supplementary~\ref{supp-sec:supp_results}.

\noindent\textbf{Continued \five{} adaptation.} We compare regional distillation with contrastive fine-tuning to test whether further training on \five{} patches can account for the gains from \methodname{}. Continued \five{} training adapts pretrained UNI2-h with the SimCLR objective \cite{pmlr-v119-chen20j}. Table~\ref{tab:contrastive_finetuning} shows that our \methodname{} achieves higher means than continued \five{} adaptation on all three tasks. Continued \five{} adaptation is close to native \five{} on IMP and TCGA-STAD, but has lower AUROC on CAM16. Additional training on these \five{} regions does not reproduce the gains from \methodname{} under the tested contrastive objective. Supp.\ref{supp-sec:supp_results} provides further details.

\begin{table}[!htbp]
\centering
\footnotesize
\setlength{\tabcolsep}{2pt}
\begin{tabular*}{\columnwidth}{@{\extracolsep{\fill}}lccc@{}}
\toprule
Encoder & CAM16 & IMP & TCGA-STAD \\
 & AUROC & Macro-F1 & C-index \\
\midrule
Native \five{} & $0.9510_{\pm0.0490}$ & $0.8982_{\pm0.0083}$ & $0.6287_{\pm0.0554}$ \\
Native \twenty{} & $\mathbf{0.9961}_{\pm0.0088}$ & $0.8970_{\pm0.0108}$ & $0.6352_{\pm0.0466}$ \\
\shortstack[l]{Contrastive\\fine-tuning} & $0.8971_{\pm0.0443}$ & $0.9002_{\pm0.0070}$ & $0.6276_{\pm0.0524}$ \\
\midrule
\textbf{\methodname{}} & $0.9686_{\pm0.0408}$ & $\mathbf{0.9078}_{\pm0.0091}$ & $\mathbf{0.6434}_{\pm0.0518}$ \\
\bottomrule
\end{tabular*}
\caption{\textbf{.} We compare contrastive learning on \five{} patches with regional distillation from their aligned \twenty{} children. Native UNI2-h features provide references at both magnifications.}
\label{tab:contrastive_finetuning}
\end{table}

\noindent\textbf{Regional aggregation.}
We ablate how event indicators from the \twenty{} teacher are aggregated to supervise the \five{} student.
OR records every event detected in any child patch, whereas AND records only the common events shared by every child. OR minus AND keeps the unique events that appeared in each child.
Occupancy is the normalized event frequency, which instead counts how many child patches contain each detected event and divides by the number of children in the region. For example, if a latent activates in 4 of 16 children, its occupancy target is $0.25$. Table~\ref{tab:regional_aggregation} shows that OR achieves the best performance in this comparison. These results support retaining both shared and localized detections in the regional target.
OR provides this supervision by recording whether each event occurs within a region, without requiring the student to estimate how frequently it occurs across the child patches.

\begin{table}[!htbp]
\centering
\footnotesize
\setlength{\tabcolsep}{2pt}
\begin{tabular*}{\columnwidth}{@{\extracolsep{\fill}}lccc@{}}
\toprule
Design & CAM16 & IMP & TCGA-STAD \\
 & AUROC & Macro-F1 & C-index \\
\midrule
AND & $0.9451_{\pm0.0567}$ & $0.8994_{\pm0.0110}$ & $0.6361_{\pm0.0585}$ \\
OR $-$ AND & $0.9657_{\pm0.0449}$ & $0.9063_{\pm0.0105}$ & $0.6359_{\pm0.0520}$ \\
Occupancy & $0.9520_{\pm0.0595}$ & $0.9018_{\pm0.0103}$ & $0.6336_{\pm0.0575}$ \\
\midrule
\textbf{OR} & $\mathbf{0.9686}_{\pm0.0408}$ & $\mathbf{0.9078}_{\pm0.0091}$ & $\mathbf{0.6434}_{\pm0.0518}$ \\
\bottomrule
\end{tabular*}
\caption{\textbf{Regional aggregation ablation.} We compare OR, AND, OR $-$ AND and occupancy targets from the same child codes. Occupancy records the fraction of children activating each latent.}
\label{tab:regional_aggregation}
\end{table}

\noindent\textbf{Native-feature anchoring.} We study whether preserving the pretrained \five{} representation improves downstream performance after distillation. An anchor loss constrains the changes to the student's weights or feature representations relative to the pretrained encoder. Parameter anchoring follows L2-SP~\cite{pmlr-v80-li18a} and penalizes changes to the pretrained weights, while relational anchoring~\cite{Park_2019_CVPR} constrains relative distances between patch features. Squared feature distance penalizes changes in both direction and magnitude, whereas the cosine anchor in Eq.~\eqref{eq:feature-anchor} constrains direction alone. Table~\ref{tab:anchor_types} shows that cosine anchoring improves over all other designs except for relational anchor on CAM16, while their difference is small. These results support preserving native feature directions during distillation. The cosine anchor constrains downstream features while allowing their magnitudes to adapt to the teacher's regional targets.

\begin{table}[!htbp]
\centering
\footnotesize
\setlength{\tabcolsep}{2pt}
\begin{tabular*}{\columnwidth}{@{\extracolsep{\fill}}lccc@{}}
\toprule
Design & CAM16 & IMP & TCGA-STAD \\
 & AUROC & Macro-F1 & C-index \\
\midrule
No anchor & $0.9569_{\pm0.0429}$ & $0.8973_{\pm0.0087}$ & $0.6285_{\pm0.0220}$ \\
Parameters & $0.9324_{\pm0.0342}$ & $0.9038_{\pm0.0104}$ & $0.6283_{\pm0.0494}$ \\
Relations & $\mathbf{0.9696}_{\pm0.0340}$ & $0.8998_{\pm0.0112}$ & $0.6148_{\pm0.0572}$ \\
Squared L2 & $0.9676_{\pm0.0423}$ & $0.9041_{\pm0.0054}$ & $0.6232_{\pm0.0517}$ \\
\midrule
\textbf{Cosine} & $0.9686_{\pm0.0408}$ & $\mathbf{0.9078}_{\pm0.0091}$ & $\mathbf{0.6434}_{\pm0.0518}$ \\
\bottomrule
\end{tabular*}
\caption{\textbf{Native-feature anchoring ablation.} We compare no anchor with parameter, relational, squared L2 and cosine penalties. Squared L2 constrains both feature direction and magnitude, while cosine constrains direction alone.}
\label{tab:anchor_types}
\end{table}

\begin{table}[!htbp]
\centering
\footnotesize
\setlength{\tabcolsep}{2pt}
\begin{tabular*}{\columnwidth}{@{\extracolsep{\fill}}lccc}
\toprule
& \multicolumn{2}{c}{Native} & \methodname{} (ours) \\
\cmidrule(lr){2-3}
MIL & \five{} & \twenty{} & \five{} \\
\midrule
\multicolumn{4}{l}{CAM16 (AUROC)} \\
TransMIL & \cellcolor{resultSecond}$0.9046_{\pm0.0422}$ & $0.8073_{\pm0.0596}$ & \cellcolor{resultBest}$\mathbf{0.9086}_{\pm0.0561}$ \\
nnMIL & \cellcolor{resultSecond}$0.9502_{\pm0.0270}$ & \cellcolor{resultBest}$\mathbf{0.9608}_{\pm0.0246}$ & $0.9497_{\pm0.0227}$ \\
RRT-MIL & $0.9270_{\pm0.0577}$ & \cellcolor{resultBest}$\mathbf{0.9322}_{\pm0.0440}$ & \cellcolor{resultSecond}$0.9281_{\pm0.0539}$ \\
\midrule
\multicolumn{4}{l}{Private-CRC (Macro-F1)} \\
TransMIL & \cellcolor{resultSecond}$0.8428_{\pm0.0257}$ & $0.8407_{\pm0.0128}$ & \cellcolor{resultBest}$\mathbf{0.8561}_{\pm0.0106}$ \\
nnMIL & \cellcolor{resultSecond}$0.8234_{\pm0.0193}$ & $0.8049_{\pm0.0111}$ & \cellcolor{resultBest}$\mathbf{0.8261}_{\pm0.0399}$ \\
RRT-MIL & $0.8199_{\pm0.0211}$ & \cellcolor{resultBest}$\mathbf{0.8362}_{\pm0.0168}$ & \cellcolor{resultSecond}$0.8288_{\pm0.0219}$ \\
\midrule
\multicolumn{4}{l}{PANDA (QW $\kappa$)} \\
TransMIL & $0.8252_{\pm0.0069}$ & \cellcolor{resultBest}$\mathbf{0.8581}_{\pm0.0101}$ & \cellcolor{resultSecond}$0.8497_{\pm0.0105}$ \\
nnMIL & $0.8395_{\pm0.0025}$ & \cellcolor{resultBest}$\mathbf{0.8675}_{\pm0.0050}$ & \cellcolor{resultSecond}$0.8556_{\pm0.0031}$ \\
RRT-MIL & $0.8340_{\pm0.0068}$ & \cellcolor{resultBest}$\mathbf{0.8632}_{\pm0.0065}$ & \cellcolor{resultSecond}$0.8604_{\pm0.0056}$ \\
\midrule
\multicolumn{4}{l}{TCGA-LUAD (C-index)} \\
TransMIL & \cellcolor{resultSecond}$0.5719_{\pm0.0346}$ & \cellcolor{resultBest}$\mathbf{0.5904}_{\pm0.0380}$ & $0.5648_{\pm0.0274}$ \\
nnMIL & $0.4953_{\pm0.0316}$ & \cellcolor{resultBest}$\mathbf{0.5653}_{\pm0.0537}$ & \cellcolor{resultSecond}$0.5507_{\pm0.0317}$ \\
RRT-MIL & \cellcolor{resultSecond}$0.5613_{\pm0.0317}$ & $0.5434_{\pm0.0181}$ & \cellcolor{resultBest}$\mathbf{0.5748}_{\pm0.0445}$ \\
\midrule
\multicolumn{4}{l}{TCGA-STAD (C-index)} \\
TransMIL & $0.5400_{\pm0.0505}$ & \cellcolor{resultBest}$\mathbf{0.5656}_{\pm0.0206}$ & \cellcolor{resultSecond}$0.5436_{\pm0.0495}$ \\
nnMIL & \cellcolor{resultBest}$\mathbf{0.5716}_{\pm0.0478}$ & $0.5408_{\pm0.0857}$ & \cellcolor{resultSecond}$0.5507_{\pm0.0175}$ \\
RRT-MIL & $0.5331_{\pm0.0324}$ & \cellcolor{resultBest}$\mathbf{0.5779}_{\pm0.0257}$ & \cellcolor{resultSecond}$0.5402_{\pm0.0380}$ \\
\midrule
\multicolumn{4}{l}{TCGA-UCEC (C-index)} \\
TransMIL & \cellcolor{resultSecond}$0.8267_{\pm0.0549}$ & $0.7525_{\pm0.0413}$ & \cellcolor{resultBest}$\mathbf{0.8287}_{\pm0.0559}$ \\
nnMIL & \cellcolor{resultBest}$\mathbf{0.7109}_{\pm0.0635}$ & $0.7059_{\pm0.1180}$ & \cellcolor{resultSecond}$0.7089_{\pm0.0450}$ \\
RRT-MIL & \cellcolor{resultBest}$\mathbf{0.8376}_{\pm0.0150}$ & $0.7931_{\pm0.0662}$ & \cellcolor{resultSecond}$0.8297_{\pm0.0217}$ \\
\bottomrule
\end{tabular*}
\caption{\textbf{Comparison across MIL architectures.} Each MIL architecture uses UNI2-h features at \five{} and \twenty{}, or \methodname{} at \five{}.}
\label{tab:mil_comparison}
\end{table}

\noindent\textbf{Downstream architectures.} We study whether distillation improves on native UNI2-h features with different MIL architectures using TransMIL~\cite{shao2021transmil}, nnMIL~\cite{luo2026nnmil} and RRT-MIL~\cite{tang2024rrt}. Table~\ref{tab:mil_comparison} shows that \methodname{} generally improves classification and grading over native \five{} across the three MIL architectures, while survival gains vary by dataset and architecture. These results suggest that regional occurrence supervision can benefit different MIL models.

%% file: main/10_conclusion.tex
\section{Conclusion and Limitations}
\label{sec:conclusion}

We introduced \methodname{}, a novel cross-magnification distillation framework that fine-tunes an encoder to predict regional events from aligned high-magnification patches using low-magnification patches as inputs. Combining sparse coding with logical OR pooling retains both localized and shared detections in the regional targets, while the dictionary provides a way to inspect the tissue patterns underlying the supervision. During distillation, a temporary prediction head maps the student's \five{} features to event occurrence probabilities, while a cosine anchor constrains changes to the pretrained feature directions. Neither SAE training nor distillation requires slide labels, which enables the large-scale training of efficient patch encoders. At inference, the adapted encoder extracts \five{} patch embeddings for downstream prediction.
We evaluated \methodname{} on 12 clinical cohorts, demonstrating strong classification and grading performance and improved survival prediction over native \five{}.

These findings motivate broader evaluation of regional event supervision and further reductions in the resources required for adaptation and inference. First, we use UNI2-h with a TopK SAE for transfer from \twenty{} to \five{}. Evaluating alternative combinations across magnification pairs and encoders would clarify whether the gains extend beyond the current configuration. In addition, survival gains vary across MIL architectures. Studying how the distilled patch features interact with aggregation and survival optimization could help explain this variation and identify settings with more consistent gains. Moreover, the student encoder is a fully fine-tuned UNI2-h architecture, so other architectures or methods with low fine-tuning cost could be explored. These extensions could broaden the applicability of regional event supervision in WSI analysis.

%% file: main/11_llm_usage.tex
\section*{LLM Usage Statement}
We used OpenAI GPT-6-Astra to assist with manuscript polishing, grammar checks, code implementations and debugging. For manuscript polishing and grammar checks, all sections except for the introduction are reviewed and modified by the LLM, with human proofreading afterwards. For code implementation and debugging, the LLM assisted in implementing the core SRED features, with human review afterwards. We are responsible for the experiments, citations, and interpretations of visual results.

%% file: supplementary/A_implementation.tex
\section{Implementation Details}
\label{sec:supp_implementation}

\subsection{Image preprocessing}

We extract $256\times256$ tissue patches from WSIs at \five{} and \twenty{} using AtlasPatch~\cite{atlaspatch}. Each \five{} patch is paired with tissue patches from the $4\times4$ grid covering the same region at \twenty{}. For UNI2-h~\cite{chen2024uni}, we resize the RGB patches to $224\times224$ with bilinear interpolation, scale pixel values to $[0,1]$, and apply the encoder's standard channel normalization.

\subsection{SAE training}

We adopt TopK~\cite{gao2025scaling} with the SAE training configuration in Table~\ref{tab:supp_sae_training}. We initialize the decoder with random vectors of Euclidean length one, copy its transpose to the encoder, and set both biases to zero. Before each optimizer step, we project decoder gradients perpendicular to the dictionary vectors. Each vector is rescaled to length one after the update. We shuffle training features without dropping the final smaller batch. The dictionary detects events for distillation.

\begin{table}[!htbp]
\centering
\small
\begin{tabular}{@{}ll@{}}
\toprule
Setting & Value \\
\midrule
Dictionary size $d_{\mathrm{SAE}}$ & 16,384 \\
TopK sparsity $k$ & 32 \\
Optimizer & Adam \\
Learning rate & $10^{-3}$, constant \\
Adam $(\beta_1,\beta_2)$ & $(0.9,0.999)$ \\
Adam $\epsilon$ & $10^{-8}$ \\
Weight decay & 0 \\
Batch size & 16,384 \\
Optimizer updates & 31,102 \\
Auxiliary sparsity $k_{\mathrm{aux}}$ & 256 \\
Auxiliary weight $\lambda_{\mathrm{aux}}$ & $1/32$ \\
Inactivity window & $2^{20}$ training examples \\
Numerical precision & FP32 \\
\bottomrule
\end{tabular}
\caption{SAE training configuration. The inactivity window determines eligibility for auxiliary reconstruction.}
\label{tab:supp_sae_training}
\end{table}

\subsection{Distillation training}

The student starts from the same pretrained UNI2-h weights as the teacher and native baselines. Its MLP head has 1,024 hidden units, GELU~\cite{hendrycks2016gelu}, and $d_{\mathrm{SAE}}$ output logits. Each output bias is initialized from the event's prevalence in the training targets. During 512 warm-up updates, the MLP learns to predict regional events from pretrained features using $\mathcal L_{\mathrm{occ}}$, while the encoder remains frozen. We jointly optimize the encoder and MLP with $\mathcal L_{\mathrm{KD}}$ on shuffled training regions. The encoder provides downstream embeddings.

Table~\ref{tab:supp_kd_training} summarizes the optimizer and schedule. A single optimizer spans the MLP warm-up and joint training. Learning rates increase linearly over the first 5\% of all updates and then decay to zero with a cosine schedule. Weight decay applies to weight matrices, with zero decay for biases and one-dimensional parameters. Gradient accumulation gives an effective batch size of 64, and each loss is normalized by the actual number of examples in the batch. The student runs in evaluation mode with gradients enabled during joint optimization. This keeps stochastic encoder operations disabled while its weights are updated.

\begin{table}[!htbp]
\centering
\small
\begin{tabular}{@{}ll@{}}
\toprule
Setting & Value \\
\midrule
Student initialization & Pretrained UNI2-h \\
MLP hidden dimension & 1,024 \\
Optimizer & AdamW \\
Encoder learning rate & $10^{-5}$ \\
MLP learning rate & $10^{-4}$ \\
AdamW $(\beta_1,\beta_2)$ & $(0.9,0.999)$ \\
AdamW $\epsilon$ & $10^{-10}$ \\
Weight decay & 0.01 \\
Effective batch / microbatch & 64 / 16 \\
MLP warm-up & 512 updates \\
Total optimizer updates & 177,292 \\
Learning-rate warm-up & First 5\% of updates \\
Learning-rate decay & Cosine \\
Gradient norm limit & 1 \\
Anchor strength $\lambda_F$ & 0.004 \\
Numerical precision & FP32 \\
\bottomrule
\end{tabular}
\caption{Training config for cross-magnification distillation.}
\label{tab:supp_kd_training}
\end{table}

The main KD ablations use the same architecture and optimizer on 751,483 training regions. Each run starts from the pretrained weights, uses the same sampling seed, and trains for 47,480 updates, including 512 MLP warm-up updates before joint optimization.

\subsection{Downstream models}

Each slide is represented by a bag of patch embeddings for MIL training. We standardize each feature dimension using the mean and standard deviation estimated from the training partition. ABMIL~\cite{ilse2018attention} first projects each embedding to 512 dimensions with ReLU, then computes its weight through gated attention with 128 hidden units. Classification and grading use cross-entropy, while survival uses four discrete hazards with equal total training weight per patient.

All shared MIL comparisons use AdamW with learning rate $2\times10^{-4}$, weight decay $10^{-4}$, effective batch size 16 and microbatch size one. For a fair comparison, ABMIL, TransMIL~\cite{shao2021transmil}, nnMIL~\cite{luo2026nnmil} and RRT-MIL~\cite{tang2024rrt} start with the same 30-epoch training budget. A new validation maximum at the final epoch indicates that further training may help. We extend these runs by 20 epochs, then by 50 epochs whenever the same condition holds at the next boundary. All reported runs finished within 100 epochs. Checkpoints are selected over the full training history, with earlier epochs preferred on exact ties. For nnMIL, we use its aggregation architecture under this protocol.

For IMP and IMP-Cervix, we average the patch embeddings of each slide and evaluate linear probes using L2-regularized logistic regression. Features are standardized using the training partition, and validation selects $C\in\{0.01,0.1,1,10,100\}$. We evaluate the probes on the common development evaluation set for each dataset.

\subsection{External baselines}

XMAG~\cite{su2026xmag} uses its DINOv2~\cite{oquab2023dinov2} ViT-B/14 student, its 224-pixel input preprocessing and 768-dimensional embeddings. For slide preparation, we use all exterior tissue regions identified by SAM2~\cite{ravi2025sam2} and select patches within their contours. This gives XMAG its own sampling grid and physical field of view. The main comparison marks datasets included in XMAG's published pretraining corpus.

We implement LRMIL~\cite{shin2026lrmil} following its patch and slide distillation stages. Patch distillation matches layers 0, 1 and 11 between aligned \five{} and \twenty{} inputs. Following the settings reported in LRMIL, we use Adam at $10^{-4}$ and slide loss weights of 0.5, 0.5 and 0.3 for bag, soft-attention and hard-attention supervision, respectively. For our experiments, patch distillation uses ten passes with batch size 16, and both attention temperatures are set to one. The adapted patch encoder supplies the inputs for slide distillation. Teacher MIL training has a maximum of 30 epochs. Both teacher and student MIL models use the paper's validation patience of 20 epochs, with no fixed epoch limit for the student. For survival, we adapt LRMIL to the four-bin censored hazard likelihood and use KL divergence between teacher and student event probabilities. For the two linear-probe datasets, patch distillation is followed by mean pooling and logistic regression. LRMIL uses its own trained prediction model throughout the MIL comparisons.

%% file: supplementary/B_method_details.tex
\section{Method Details}
\label{sec:supp_method}

\subsection{SAE input normalization}
\label{sec:supp_feature_normalization}

We apply z-score normalization with a shared scale factor to center teacher features and place SAE inputs on a common numerical scale. Given $M$ training features $t_r\in\mathbb R^d$, the mean and scale are
\begin{equation}
  \mu=\frac{1}{M}\sum_{r=1}^{M}t_r,
  \qquad
  s^2=\frac{1}{Md}\sum_{r=1}^{M}\|t_r-\mu\|_2^2,
  \label{eq:feature-normalization}
\end{equation}
and the SAE input is $\widetilde t=(t-\mu)/s$. These statistics are estimated from the SAE training features and fixed throughout dictionary training and regional target construction.

Centering expresses each input as a deviation from the average teacher representation. The shared scale gives the training features unit average coordinate variance. Thus, reconstruction error is measured relative to the total variation in the training inputs. Scaling all coordinates equally preserves relative feature distances, variation among coordinates, and the relative lengths of centered vectors.

%% file: supplementary/C_datasets.tex
\section{Datasets}
\label{sec:supp_data}

\subsection{SAE training data}
We train the SAE on 254,780,319 \twenty{} features from 24,556 slides spanning 24 anatomical groups. Features are sampled uniformly from the combined datasets in Table~\ref{tab:sae_cohorts}.

\begin{table}[!htbp]
\centering
\small
\setlength{\tabcolsep}{2pt}
\begin{tabular*}{\columnwidth}{@{\extracolsep{\fill}}lrr@{}}
\toprule
Dataset & Slides & \twenty{} features \\
\midrule
TCGA & 9,501 & 132,633,661 \\
CPTAC & 6,738 & 35,702,601 \\
EBRAINS & 3,114 & 33,283,668 \\
MUT-HET-RCC & 781 & 8,679,542 \\
OV-Bevacizumab & 286 & 5,608,897 \\
BC-therapy & 166 & 340,677 \\
NADT-Prostate & 490 & 449,010 \\
BOEHMK, H\&E/MSKCC subset & 159 & 1,197,901 \\
MBC & 99 & 559,982 \\
\midrule
In-house IPMN & 288 & 6,103,465 \\
In-house colorectal & 177 & 3,606,386 \\
In-house PDAC biopsies & 1,005 & 1,919,932 \\
In-house pancreatic collection & 1,752 & 24,694,597 \\
\midrule
Total & 24,556 & 254,780,319 \\
\bottomrule
\end{tabular*}
\caption{\textbf{SAE training datasets.} Training slides and features after dataset exclusions and removal of duplicate coordinates.}
\label{tab:sae_cohorts}
\end{table}

\noindent\textbf{TCGA and CPTAC.}
TCGA~\cite{tcga_resource,heath2021gdc} contributes 27 cancer cohorts: ACC, BLCA, BRCA, CESC, CHOL, COAD, DLBC, ESCA, GBM, HNSC, KICH, LGG, LIHC, LUSC, MESO, OV, PAAD, PCPG, PRAD, READ, SARC, SKCM, TGCT, THCA, THYM, UCS and UVM. The five TCGA datasets used for downstream survival evaluation are excluded. CPTAC~\cite{li2023cptac} adds ten collections covering breast, kidney, colon, brain, head and neck, lung, ovarian, pancreatic and endometrial cancers.

\noindent\textbf{EBRAINS.}
The digital brain tumour atlas~\cite{roetzer2022ebrains} provides diverse central nervous system pathology. Our training subset contains 3,067 tumour slides representing 126 diagnosis labels and 47 control slides.

\noindent\textbf{Renal and prostate datasets.}
MUT-HET-RCC~\cite{acosta2022heterogeneity} examines intratumour heterogeneity in renal cell carcinoma. NADT-Prostate~\cite{wilkinson2020nadt} contains prostate cancer histology for neoadjuvant androgen deprivation therapy studies.

\noindent\textbf{Ovarian and breast datasets.}
OV-Bevacizumab~\cite{wang2022ovarian,wang2021ovariandata} contains ovarian cancer histology associated with bevacizumab treatment response. The H\&E/MSKCC subset of BOEHMK~\cite{boehm2022multimodal} contributes high-grade serous ovarian cancer slides from a multimodal risk stratification study. BC-therapy~\cite{sammut2022therapy,sammut2022therapydata} provides breast cancer biopsies from a study of response to neoadjuvant therapy. MBC~\cite{bergstrom2024hrd,galland2022platinum} adds metastatic breast cancer samples studied for homologous recombination deficiency and platinum response.

\noindent\textbf{In-house datasets.}
Our in-house datasets comprise intraductal papillary mucinous neoplasms (IPMN), colorectal specimens, pancreatic ductal adenocarcinoma (PDAC) biopsies and a separate pancreatic collection. The latter provides 24,694,597 SAE training features from 1,752 slides. Distillation uses collections with slide images.

\subsection{Cross-magnification distillation}
We train the student on aligned \five{} regions and their \twenty{} children from the same 24 anatomical groups. Public datasets use their predefined partitions. Within each remaining dataset and anatomical group, we assign patient or specimen groups to training (70\%), validation (15\%) and evaluation (15\%). Slides in each group stay together (Table~\ref{tab:source_kd_splits}).

\begin{table}[!htbp]
\centering
\small
\setlength{\tabcolsep}{3pt}
\begin{tabular*}{\columnwidth}{@{\extracolsep{\fill}}lrrr@{}}
\toprule
Partition & Slides & Groups & \five{} regions \\
\midrule
Training & 15,963 & 10,392 & 11,313,866 \\
Validation & 3,486 & 2,227 & 2,496,874 \\
Evaluation & 3,940 & 2,619 & 2,919,381 \\
\bottomrule
\end{tabular*}
\caption{\textbf{Cross-magnification distillation partitions.} Groups follow the recorded patient identifiers or the dataset's declared specimen and slide grouping rules. Linked slides share a partition.}
\label{tab:source_kd_splits}
\end{table}

\subsection{Downstream datasets}
Downstream evaluation covers seven classification or grading datasets and five survival cohorts. These datasets are excluded from the corpus used to train the SAE for distillation and from student fine-tuning.

\noindent\textbf{CAM16.}
CAMELYON16~\cite{litjens2018camelyon} evaluates breast cancer metastasis detection in lymph nodes. The 270 labeled development slides comprise 159 normal and 111 tumour slides. Partitions are defined by slide.

\noindent\textbf{CAM17.}
CAMELYON17~\cite{litjens2018camelyon} extends lymph node assessment to multiple slides per challenge group. Our 499 slides form 100 groups, with 318 negative slides, 36 with isolated tumour cells, 59 with micrometastases and 86 with macrometastases. Slides in each group share a partition.

\noindent\textbf{Private-CRC.}
This in-house colorectal polyp dataset contains 837 slides: 212 hyperplastic polyps, 201 sessile serrated lesions, 207 tubular adenomas and 217 tubulovillous adenomas. We use 757 slides for training, validation and evaluation, setting aside the original 80 test slides.

\noindent\textbf{BRACS.}
BRACS~\cite{brancati2022bracs} evaluates breast lesion classification. We use 546 slides from 189 patients, including 44 normal, 147 pathological benign, 73 usual ductal hyperplasia, 41 flat epithelial atypia, 48 atypical ductal hyperplasia, 61 ductal carcinoma in situ and 132 invasive carcinoma slides. We construct partitions by patient using the slide labels.

\noindent\textbf{PANDA.}
PANDA~\cite{bulten2022panda} evaluates prostate biopsy grading. The 10,615 slides comprise 2,891, 2,666, 1,343, 1,242, 1,249 and 1,224 examples of ISUP grades 0--5, respectively. We stratify the slide partitions by grade.

\noindent\textbf{IMP.}
IMP~\cite{neto2024impcrs} contains 5,333 colorectal biopsy and polypectomy slides from IMP Diagnostics. The three classes are non-neoplastic tissue, low-grade lesions and high-grade lesions. CRS1 and CRS2~\cite{oliveira2021cad,neto2022imil4path} provide the 4,433 development slides, and the 900-slide CRS-Test collection~\cite{neto2024interpretable} is set aside.

\noindent\textbf{IMP-Cervix.}
IMP-Cervix~\cite{oliveira2024impcervix,oliveira2023cervical} contains cervical excision specimens from IMP Diagnostics, labeled as non-neoplastic, low-grade squamous intraepithelial lesion, high-grade squamous intraepithelial lesion or non-representative. We divide the released 600-slide set into 360 training, 91 validation and 149 development evaluation slides.

\noindent\textbf{TCGA survival datasets.}
Overall survival evaluation uses kidney renal clear cell carcinoma (KIRC), kidney renal papillary cell carcinoma (KIRP), lung adenocarcinoma (LUAD), stomach adenocarcinoma (STAD) and uterine corpus endometrial carcinoma (UCEC)~\cite{tcga_resource}. Their labeled inventories contain 517, 296, 528, 409 and 275 slides from 511, 272, 465, 383 and 249 cases, respectively. Survival endpoints follow the TCGA clinical framework~\cite{liu2018clinical}. We partition these datasets by case and average slide predictions for cases with multiple slides.

%% file: supplementary/D_additional_results.tex
\begin{table*}[tbp]
\centering
\footnotesize
\begin{tabular*}{\textwidth}{@{\extracolsep{\fill}}llccccc@{}}
\toprule
& & \multicolumn{2}{c}{Native} & \multicolumn{3}{c}{Cross-magnification KD methods} \\
\cmidrule(lr){3-4}\cmidrule(lr){5-7}
Dataset & Metric & \five{} & \twenty{} & XMAG~\cite{su2026xmag} & LRMIL~\cite{shin2026lrmil} & \methodname{} (ours) \\
\midrule
CAM16 & Accuracy $\uparrow$ & \cellcolor{resultThird}$0.9306_{\pm0.0354}$ & \cellcolor{resultBest}$\mathbf{0.9611}_{\pm0.0206}$ & $0.8722_{\pm0.0248}$ & $0.8000_{\pm0.0990}$ & \cellcolor{resultSecond}$\underline{0.9528}_{\pm0.0158}$ \\
 & AUROC $\uparrow$ & \cellcolor{resultSecond}$0.9773_{\pm0.0146}$ & \cellcolor{resultThird}$0.9719_{\pm0.0247}$ & $0.9275_{\pm0.0195}$ & $0.8897_{\pm0.0378}$ & \cellcolor{resultBest}$\underline{\mathbf{0.9905}}_{\pm0.0105}$ \\
 & Macro-F1 $\uparrow$ & \cellcolor{resultThird}$0.9288_{\pm0.0359}$ & \cellcolor{resultBest}$\mathbf{0.9594}_{\pm0.0216}$ & $0.8673_{\pm0.0269}$ & $0.7687_{\pm0.1563}$ & \cellcolor{resultSecond}$\underline{0.9514}_{\pm0.0159}$ \\
\midrule
CAM17 & Accuracy $\uparrow$ & \cellcolor{resultThird}$0.8292_{\pm0.0453}$ & \cellcolor{resultSecond}$0.8462_{\pm0.0308}$ & $0.7846_{\pm0.1042}$ & \cellcolor{resultBest}$\underline{\mathbf{0.8523}}_{\pm0.0167}$ & $0.8154_{\pm0.0326}$ \\
 & AUROC $\uparrow$ & \cellcolor{resultThird}$0.8515_{\pm0.0149}$ & \cellcolor{resultSecond}$0.8701_{\pm0.0174}$ & $0.8238_{\pm0.0293}$ & $0.7989_{\pm0.0293}$ & \cellcolor{resultBest}$\underline{\mathbf{0.8792}}_{\pm0.0100}$ \\
 & Macro-F1 $\uparrow$ & \cellcolor{resultThird}$0.5894_{\pm0.0310}$ & \cellcolor{resultBest}$\mathbf{0.6705}_{\pm0.0375}$ & $0.5508_{\pm0.0535}$ & $0.5118_{\pm0.0318}$ & \cellcolor{resultSecond}$\underline{0.6116}_{\pm0.0515}$ \\
\midrule
Private-CRC & Accuracy $\uparrow$ & $0.8138_{\pm0.0115}$ & \cellcolor{resultBest}$\mathbf{0.8265}_{\pm0.0177}$ & \cellcolor{resultThird}$0.8212_{\pm0.0210}$ & $0.7831_{\pm0.0351}$ & \cellcolor{resultBest}$\underline{\mathbf{0.8265}}_{\pm0.0185}$ \\
 & AUROC $\uparrow$ & \cellcolor{resultThird}$0.9609_{\pm0.0102}$ & \cellcolor{resultSecond}$0.9639_{\pm0.0037}$ & $0.9600_{\pm0.0035}$ & $0.9365_{\pm0.0221}$ & \cellcolor{resultBest}$\underline{\mathbf{0.9645}}_{\pm0.0034}$ \\
 & Macro-F1 $\uparrow$ & $0.8105_{\pm0.0124}$ & \cellcolor{resultSecond}$0.8229_{\pm0.0187}$ & \cellcolor{resultThird}$0.8191_{\pm0.0204}$ & $0.7794_{\pm0.0356}$ & \cellcolor{resultBest}$\underline{\mathbf{0.8249}}_{\pm0.0188}$ \\
\midrule
BRACS & Accuracy $\uparrow$ & \cellcolor{resultBest}$\mathbf{0.5361}_{\pm0.0212}$ & $0.4612_{\pm0.0409}$ & $0.4748_{\pm0.0201}$ & \cellcolor{resultThird}$0.4803_{\pm0.0510}$ & \cellcolor{resultSecond}$\underline{0.5034}_{\pm0.0259}$ \\
 & AUROC $\uparrow$ & \cellcolor{resultSecond}$0.8494_{\pm0.0152}$ & \cellcolor{resultThird}$0.8222_{\pm0.0346}$ & $0.8174_{\pm0.0086}$ & $0.7679_{\pm0.0633}$ & \cellcolor{resultBest}$\underline{\mathbf{0.8500}}_{\pm0.0171}$ \\
 & Macro-F1 $\uparrow$ & \cellcolor{resultBest}$\mathbf{0.4375}_{\pm0.0399}$ & \cellcolor{resultThird}$0.3875_{\pm0.0533}$ & $0.3825_{\pm0.0209}$ & $0.3021_{\pm0.1228}$ & \cellcolor{resultSecond}$\underline{0.4273}_{\pm0.0246}$ \\
\midrule
PANDA & Accuracy $\uparrow$ & \cellcolor{resultThird}$0.6923_{\pm0.0053}$ & \cellcolor{resultBest}$0.7272_{\pm0.0089}$ & $0.6804_{\pm0.0047}$ & $0.6644_{\pm0.0019}$ & \cellcolor{resultBest}$\underline{\mathbf{0.7272}}_{\pm0.0073}$ \\
 & AUROC $\uparrow$ & \cellcolor{resultThird}$0.9186_{\pm0.0063}$ & \cellcolor{resultBest}$\mathbf{0.9312}_{\pm0.0030}$ & $0.9122_{\pm0.0015}$ & $0.9051_{\pm0.0012}$ & \cellcolor{resultSecond}$\underline{0.9290}_{\pm0.0054}$ \\
 & Macro-F1 $\uparrow$ & \cellcolor{resultThird}$0.6276_{\pm0.0149}$ & \cellcolor{resultSecond}$0.6626_{\pm0.0141}$ & $0.6201_{\pm0.0074}$ & $0.5939_{\pm0.0061}$ & \cellcolor{resultBest}$\underline{\mathbf{0.6725}}_{\pm0.0054}$ \\
 & QW $\kappa$ $\uparrow$ & \cellcolor{resultThird}$0.8367_{\pm0.0078}$ & \cellcolor{resultBest}$\mathbf{0.8661}_{\pm0.0046}$ & $0.8247_{\pm0.0018}$ & $0.8181_{\pm0.0037}$ & \cellcolor{resultSecond}$\underline{0.8587}_{\pm0.0062}$ \\
 & Grade MAE $\downarrow$ & \cellcolor{resultThird}$0.4596_{\pm0.0113}$ & \cellcolor{resultBest}$\mathbf{0.3996}_{\pm0.0137}$ & $0.4824_{\pm0.0101}$ & $0.5105_{\pm0.0056}$ & \cellcolor{resultSecond}$\underline{0.4082}_{\pm0.0123}$ \\
\midrule
IMP & Accuracy $\uparrow$ & \cellcolor{resultThird}$0.9025_{\pm0.0025}$ & \cellcolor{resultSecond}$0.9047_{\pm0.0045}$ & $0.8989_{\pm0.0033}$ & $0.8995_{\pm0.0039}$ & \cellcolor{resultBest}$\underline{\mathbf{0.9083}}_{\pm0.0024}$ \\
 & AUROC $\uparrow$ & \cellcolor{resultSecond}$0.9792_{\pm0.0006}$ & \cellcolor{resultThird}$0.9761_{\pm0.0009}$ & $0.9754_{\pm0.0011}$ & $0.9743_{\pm0.0008}$ & \cellcolor{resultBest}$\underline{\mathbf{0.9794}}_{\pm0.0005}$ \\
 & Macro-F1 $\uparrow$ & \cellcolor{resultThird}$0.8991_{\pm0.0019}$ & \cellcolor{resultSecond}$0.9031_{\pm0.0050}$ & $0.8937_{\pm0.0046}$ & $0.8926_{\pm0.0024}$ & \cellcolor{resultBest}$\underline{\mathbf{0.9056}}_{\pm0.0028}$ \\
\midrule
IMP-Cervix & Accuracy $\uparrow$ & \cellcolor{resultBest}$\mathbf{0.7396}_{\pm0.0286}$ & \cellcolor{resultThird}$0.7168_{\pm0.0423}$ & $0.6859_{\pm0.0309}$ & $0.6336_{\pm0.0357}$ & \cellcolor{resultSecond}$\underline{0.7315}_{\pm0.0157}$ \\
 & AUROC $\uparrow$ & \cellcolor{resultBest}$\mathbf{0.9059}_{\pm0.0148}$ & \cellcolor{resultThird}$0.8723_{\pm0.0161}$ & $0.8687_{\pm0.0209}$ & $0.8345_{\pm0.0208}$ & \cellcolor{resultSecond}$\underline{0.9010}_{\pm0.0134}$ \\
 & Macro-F1 $\uparrow$ & \cellcolor{resultBest}$\mathbf{0.7096}_{\pm0.0337}$ & \cellcolor{resultThird}$0.6471_{\pm0.0630}$ & $0.6076_{\pm0.0660}$ & $0.5607_{\pm0.0408}$ & \cellcolor{resultSecond}$\underline{0.6884}_{\pm0.0213}$ \\
\bottomrule
\end{tabular*}
\caption{\textbf{Classification and grading.} Accuracy, AUROC and macro-F1 complement the primary endpoints in the main clinical benchmark. IMP and IMP-Cervix use mean-pooled features with logistic regression. PANDA also reports quadratic weighted $\kappa$ and grade MAE.}
\label{tab:supp_clinical_probes}
\end{table*}

\section{Additional Results and Ablations}
\label{sec:supp_results}

\subsection{Detailed clinical metrics}
\label{sec:supp_benchmark_metrics}
We report additional classification metrics for the models evaluated in the main clinical benchmark. Table~\ref{tab:supp_clinical_probes} shows that \methodname{} improves AUROC over native \five{} on most datasets and consistently exceeds both KD baselines.

\subsection{Computational efficiency}
\label{sec:computational_efficiency}

We measure the time from the original WSI to the slide prediction. Table~\ref{tab:inference_efficiency} shows that our \methodname{} achieves an $11.58\times$ speedup and reduces network compute by approximately $93.4\%$ compared with native \twenty{}. These savings arise mainly from reading and encoding fewer patches at \five{}, which also reduces the number of feature vectors stored for each slide. The reduction in peak GPU memory is smaller because all three systems use the same UNI2-h backbone and encoding batch size. Our \methodname{} also closely matches native \five{} in runtime and memory, since SAE and the temporary prediction head are used only during training.

\begin{table}[!htbp]
\centering
\small
\setlength{\tabcolsep}{2pt}
\begin{tabular*}{\columnwidth}{@{\extracolsep{\fill}}lrrrr@{}}
\toprule
System & \shortstack{Time $\downarrow$\\(s/slide)} & \shortstack{Speedup\\$\uparrow$} & \shortstack{Network $\downarrow$\\(TFLOPs/slide)} & \shortstack{GPU $\downarrow$\\(GiB)} \\
\midrule
Native \twenty{} & 70.04 & 1.00 & 5,015.38 & 3.81 \\
Native \five{} & \textbf{6.03} & \textbf{11.61} & \textbf{330.89} & \textbf{3.51} \\
\methodname{} (ours) & 6.05 & 11.58 & \textbf{330.89} & \textbf{3.51} \\
\bottomrule
\end{tabular*}
\caption{\textbf{Inference cost.} Time includes slide preparation through prediction. Speedup is relative to the total native \twenty{} time. Network FLOPs are calculated for the encoder and downstream predictor using each slide's actual patch count. GPU memory is the peak PyTorch allocation.}
\label{tab:inference_efficiency}
\end{table}

\subsection{Ablation on continued \five{} adaptation}
\label{sec:supp_continued_adaptation}

We fine-tune pretrained UNI2-h on the same approximately $0.75$ million \five{} training regions used in the main KD ablations. Two augmented views of each patch form a positive pair, while views from other patches provide negatives for the SimCLR~\cite{pmlr-v119-chen20j} objective. We use a batch size of 64 and temperature $0.5$. Training uses no \twenty{} teacher, SAE targets or feature anchor.

\subsection{Ablation on anchor strength}
\label{sec:supp_anchor_strength}

\begin{table}[!htbp]
\centering
\footnotesize
\setlength{\tabcolsep}{2pt}
\begin{tabular*}{\columnwidth}{@{\extracolsep{\fill}}lccc@{}}
\toprule
$\lambda_F$ & CAM16 & IMP & TCGA-STAD \\
 & AUROC & Macro-F1 & C-index \\
\midrule
$0.001$ & $\mathbf{0.9706}_{\pm0.0384}$ & $0.9048_{\pm0.0104}$ & $0.6357_{\pm0.0557}$ \\
$0.016$ & $0.9588_{\pm0.0555}$ & $0.9041_{\pm0.0106}$ & $0.6348_{\pm0.0556}$ \\
\midrule
$\boldsymbol{0.004}$ (Default) & $0.9686_{\pm0.0408}$ & $\mathbf{0.9078}_{\pm0.0091}$ & $\mathbf{0.6434}_{\pm0.0518}$ \\
\bottomrule
\end{tabular*}
\caption{\textbf{Cosine-anchor strength ablation.} We compare the original weight $\lambda_F$ with weaker anchor $0.25\lambda_F$ and strong one $4\lambda_F$. }
\label{tab:anchor_strength}
\end{table}

We further investigate the cosine anchor weight to study how strongly native feature directions should be constrained. The coefficient $\lambda_F$ controls the strength of the cosine anchor during fine-tuning. Larger values give more emphasis on preserving the pretrained \five{} feature directions. By default, we set $\lambda_F=0.004$. We compare it with $0.25\lambda_F$ and $4\lambda_F$. Table~\ref{tab:anchor_strength} shows that no additional weight setting gives the best results on IMP and TCGA-STAD, while the weaker anchor gives slightly higher AUROC on CAM16. The stronger penalty lowers performance on all three tasks. These results show that our default $\lambda_F$ already balances original feature directions and allows the student to learn from regional supervision.

\subsection{Ablation on prediction loss}
\label{sec:supp_prediction_loss}

\begin{table}[!htbp]
\centering
\footnotesize
\setlength{\tabcolsep}{2pt}
\begin{tabular*}{\columnwidth}{@{\extracolsep{\fill}}lccc@{}}
\toprule
Design & CAM16 & IMP & TCGA-STAD \\
 & AUROC & Macro-F1 & C-index \\
\midrule
ASL & $0.9578_{\pm0.0572}$ & $0.9038_{\pm0.0098}$ & $0.6365_{\pm0.0523}$ \\
\midrule
\textbf{BCE} & $\mathbf{0.9686}_{\pm0.0408}$ & $\mathbf{0.9078}_{\pm0.0091}$ & $\mathbf{0.6434}_{\pm0.0518}$ \\
\bottomrule
\end{tabular*}
\caption{\textbf{Occurrence-loss ablation.} ASL reduces the contribution of confidently predicted negatives.}
\label{tab:prediction_loss}
\end{table}

We compare BCE with asymmetric loss (ASL)~\cite{Ridnik_2021_ICCV} to study how the loss handles the many absent events in sparse regional targets. ASL reduces the contribution of events marked absent in the target when the student already assigns them low occurrence probabilities, while the OR targets and cosine anchor are unchanged. Table~\ref{tab:prediction_loss} shows that BCE achieves higher performance, indicating that using BCE helps the student learn which teacher events occur within each region. Predicting regional occurrence requires the student to recognize detected events and avoid predicting those absent from the aligned children's codes.

%% file: supplementary/E_qualitative.tex
\begin{table*}[!tbp]
\centering
\footnotesize
\setlength{\tabcolsep}{4pt}
\begin{tabular*}{\textwidth}{@{\extracolsep{\fill}}rrp{0.74\textwidth}@{}}
\toprule
Latent & Occurrence (\%) & Pathologist's description and comments \\
\midrule
\multicolumn{3}{@{}l}{\textit{Additional tissue structures}} \\
1385 & 27.6106 & Glandular structures from different organs. \\
4668 & 46.6330 & Large glandular structures with large lumens. \\
2874 & 42.9318 & Large glandular structures on an eosinophilic background. \\
3514 & 35.7496 & Glandular and parenchymal cells. \\
2070 & 27.2946 & Well-defined clusters of epithelial cells. \\
1159 & 38.0907 & Large clusters of tumor cells admixed with stroma. \\
670 & 43.3438 & Sheets of cancer cells from different organs. \\
1635 & 40.2116 & Spindle cells of different nature. \\
12987 & 52.2648 & Brain tumor cells on a neurofibrillar background. \\
4961 & 49.6175 & Brain tumor cells heterogeneously scattered on a neurofibrillar background. \\
2654 & 47.8387 & Brain tumor cells arranged in monotonous sheets. \\
2770 & 32.6538 & Brain tumor cells in monotonous sheets with a mild fascicular arrangement. \\
3885 & 36.4821 & Brain tumor cells in monotonous sheets with focal aberrant enlarged cells. \\
\midrule
\multicolumn{3}{@{}l}{\textit{General appearance at high regional occurrence}} \\
2124 & 81.9414 & No clear histological or pathological pattern. \\
8624 & 74.5186 & Round pale structures on an eosinophilic background. \\
9904 & 77.7109 & Possibly aligned or balanced nuclei over a monotonous background. \\
\midrule
\multicolumn{3}{@{}l}{\textit{Low regional occurrence (below 1\%)}} \\
2044 & 0.001229 & Glass. \\
5229 & 0.000053 & Sheets of cancer cells that merge into each other. This could be an artifact. \\
6046 & 0.000115 & Spindled and plump cancer cells or other structures. \\
6492 & 0.002183 & Benign brain structures with activated glia. \\
8115 & 0.001617 & Pale myofibroblastic stroma. This is not a rare pattern. \\
8503 & 0.001582 & Myxoid changes with mild artifacts. \\
12939 & 0.0763 & Benign glandular structures. \\
13419 & 0.000804 & Large bundles of smooth muscle in the stroma. \\
13745 & 0.5833 & Lymphocyte-rich lymphoid tissue. \\
14016 & 0.6200 & Muscle. \\
14294 & 0.003553 & Necrosis and cell debris. \\
14488 & 0.7500 & Brain tissue with no overt cancer in the reviewer's description. \\
14690 & 0.7724 & Adipose tissue. \\
14885 & 0.2605 & Large sheets of poorly differentiated cancer. \\
15924 & 0.002934 & Immunohistochemical contamination, immune cells and adipose tissue with mild focus artifact. \\
\bottomrule
\end{tabular*}
\caption{\textbf{Additional pathologist annotations of SAE responses.} Each row describes patches activating the indicated latent. Occurrence is the percentage of \five{} distillation training regions whose OR target includes that latent.}
\label{tab:supp_pathologist_review}
\end{table*}

\section{Additional qualitative analysis}
\label{sec:supp_qualitative}

\subsection{Pathologist review of SAE responses}

We provide additional SAE response groups reviewed by a pathologist. As in Fig.~2(e) of the main paper, regional occurrence is the percentage of \five{} training regions whose OR target includes a latent. Each group contains \twenty{} patches that activate the same latent. Table~\ref{tab:supp_pathologist_review} records glandular structures, epithelial cell clusters, muscle, stromal tissue and brain morphology. The most frequent responses have broad visual associations or no clear common pattern. Groups with lower occurrence include both specific tissue structures and responses to glass or artifacts.

\begin{figure*}[!tb]
\centering
\includegraphics[width=\textwidth]{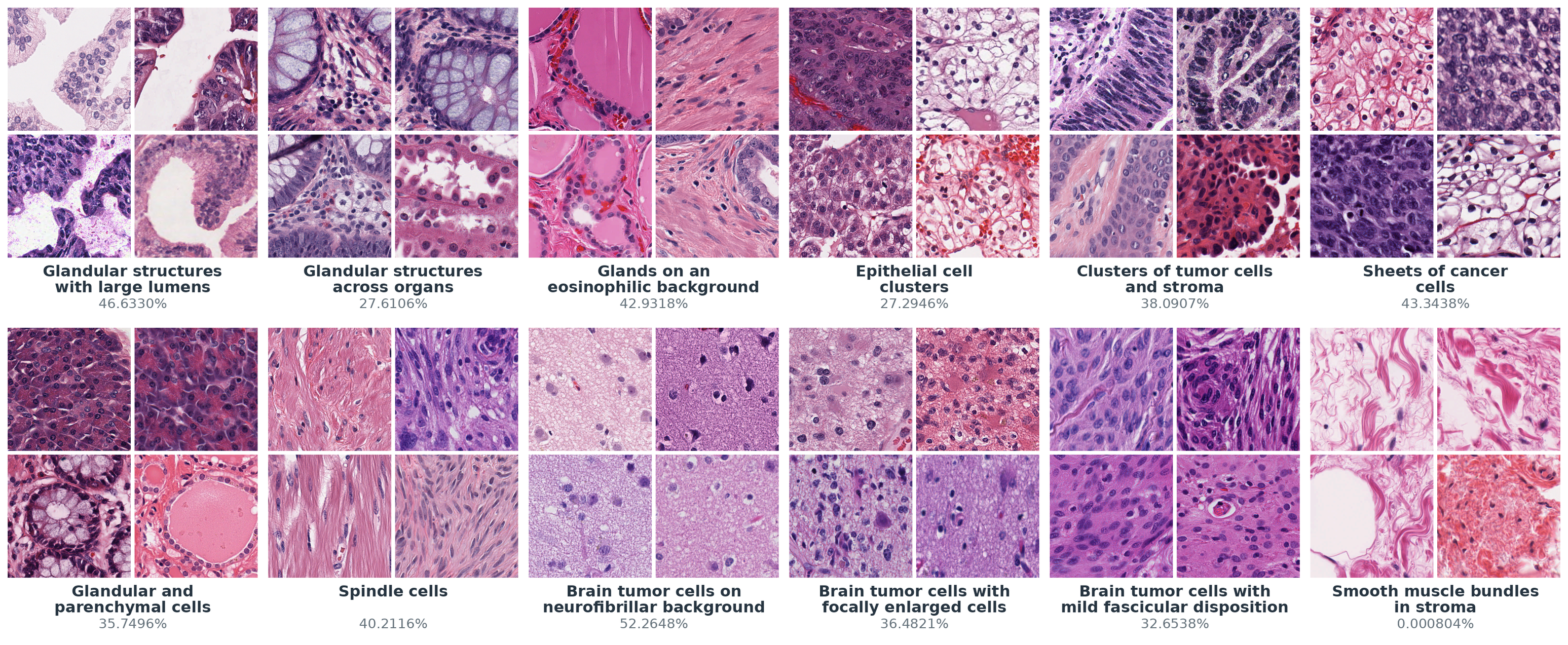}
\caption{\textbf{Additional tissue patterns associated with SAE responses.} Each $2\times2$ group contains patches nominated by the pathologist for the same latent. Labels give the tissue description and regional occurrence, as in Fig.~2 of the main paper.}
\label{fig:supp_sae_glands_cancer}
\end{figure*}

Fig.~\ref{fig:supp_sae_glands_cancer} shows patches nominated by the pathologist for glandular structures with large lumens, epithelial cell clusters, tumor cells admixed with stroma, spindle cells and smooth muscle bundles. Each $2\times2$ group shows four patches activating the same latent.

\FloatBarrier

\begin{figure*}[!tb]
\centering
\includegraphics[width=\textwidth]{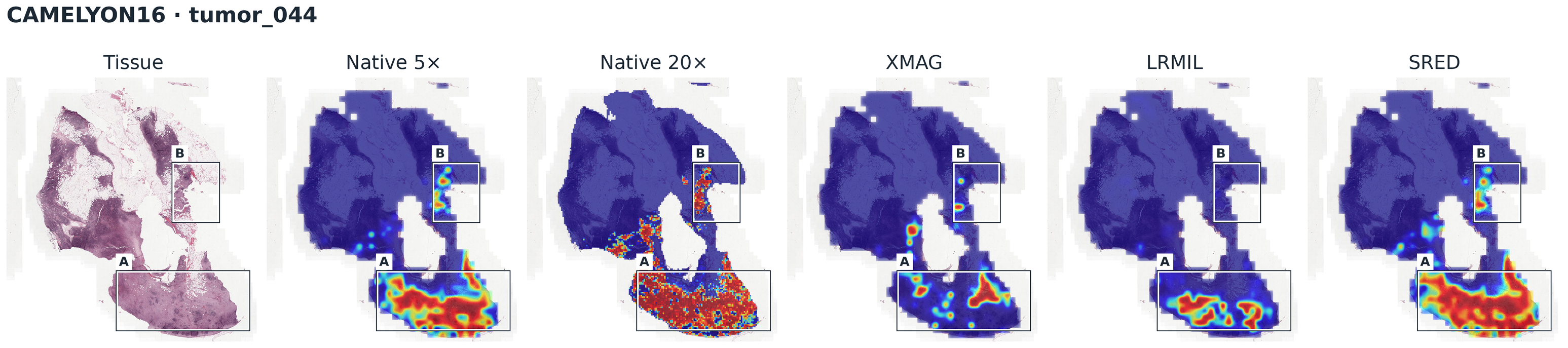}\\[3pt]
\includegraphics[width=\textwidth]{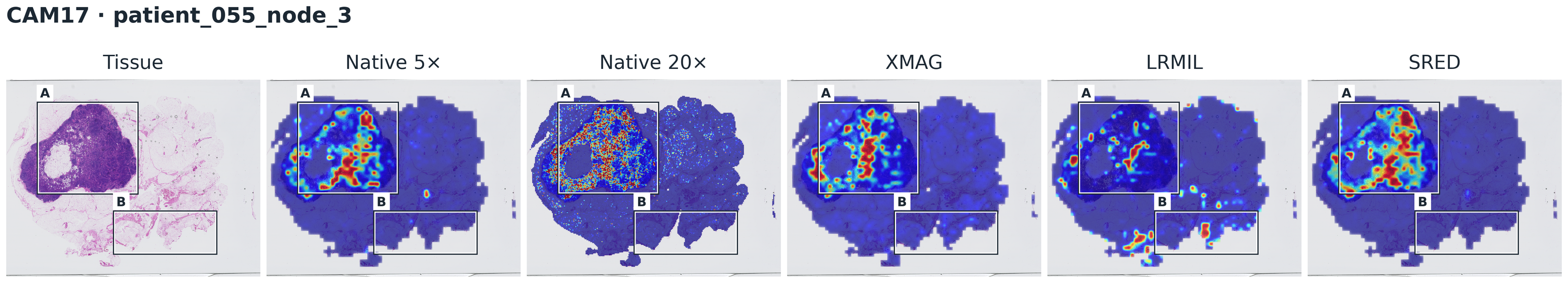}
\caption{\textbf{Additional attention comparisons for classification.} Rows show CAM16 tumor\_044 and CAM17 patient\_055\_node\_3. Both native encoders and \methodname{} emphasize the lower region on CAM16 and the densely stained left region on CAM17. A and B mark corresponding tissue locations across all panels.}
\label{fig:supp_attention_additional}
\end{figure*}

\subsection{Additional slide attention maps}

Additional attention maps from CAM16, CAM17, PANDA and three survival datasets complement the main-paper examples. Fig.~\ref{fig:supp_attention_additional} compares shared areas of attention at different spatial resolutions. On CAM16, both native encoders and \methodname{} highlight the lower tissue band in A and the right interior in B. LRMIL emphasizes A with little response in B. On CAM17, the native encoders, XMAG and \methodname{} emphasize the densely stained region A. Native \twenty{} resolves separate foci within this region, while the \five{} maps show broader responses. LRMIL has additional foci in B.

\begin{figure*}[!tb]
\centering
\includegraphics[width=\textwidth]{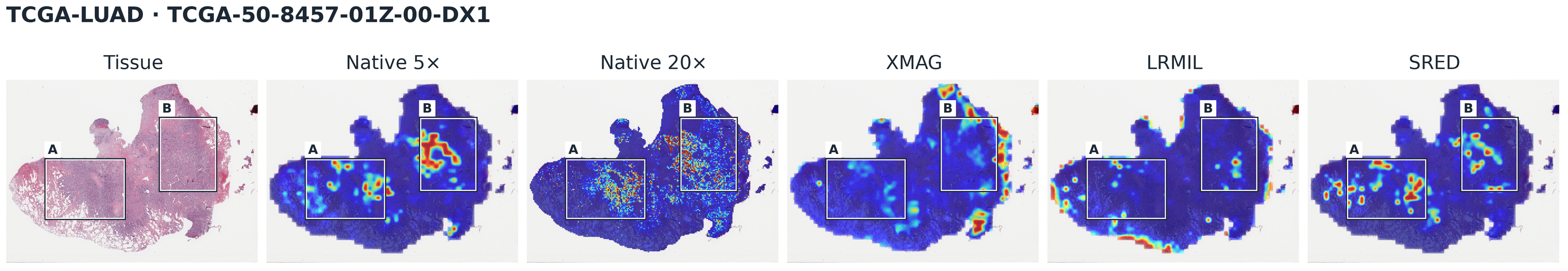}\\[3pt]
\includegraphics[width=\textwidth]{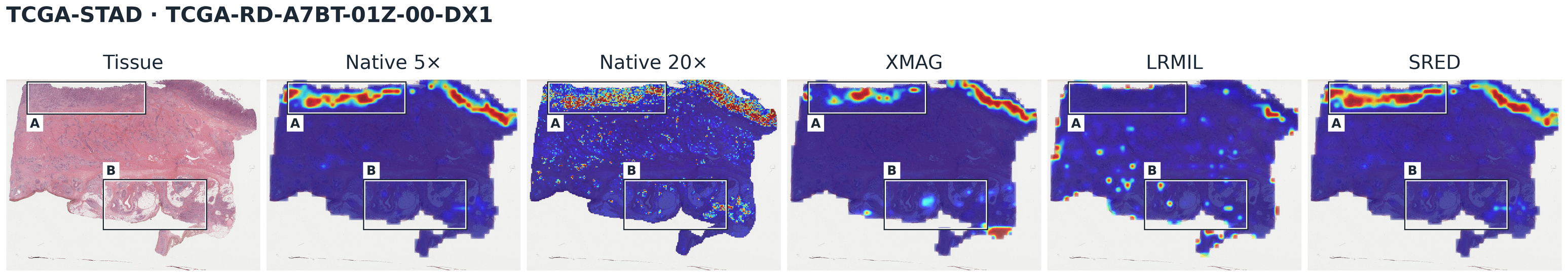}
\caption{\textbf{Attention comparisons for survival.} Rows show TCGA-LUAD TCGA-50-8457 and TCGA-STAD TCGA-RD-A7BT. A and B mark corresponding tissue locations across all panels.}
\label{fig:supp_attention_survival}
\end{figure*}

Fig.~\ref{fig:supp_attention_survival} shows lung and gastric examples. On LUAD, both native encoders and \methodname{} highlight interior regions A and B, while XMAG and LRMIL emphasize more of the tissue boundary. On STAD, the native encoders, XMAG and \methodname{} share a prominent band along the upper edge in A. LRMIL places more attention on the lower interior in B than native \five{} or \methodname{}.

\begin{figure*}[!tb]
\centering
\includegraphics[width=0.94\textwidth]{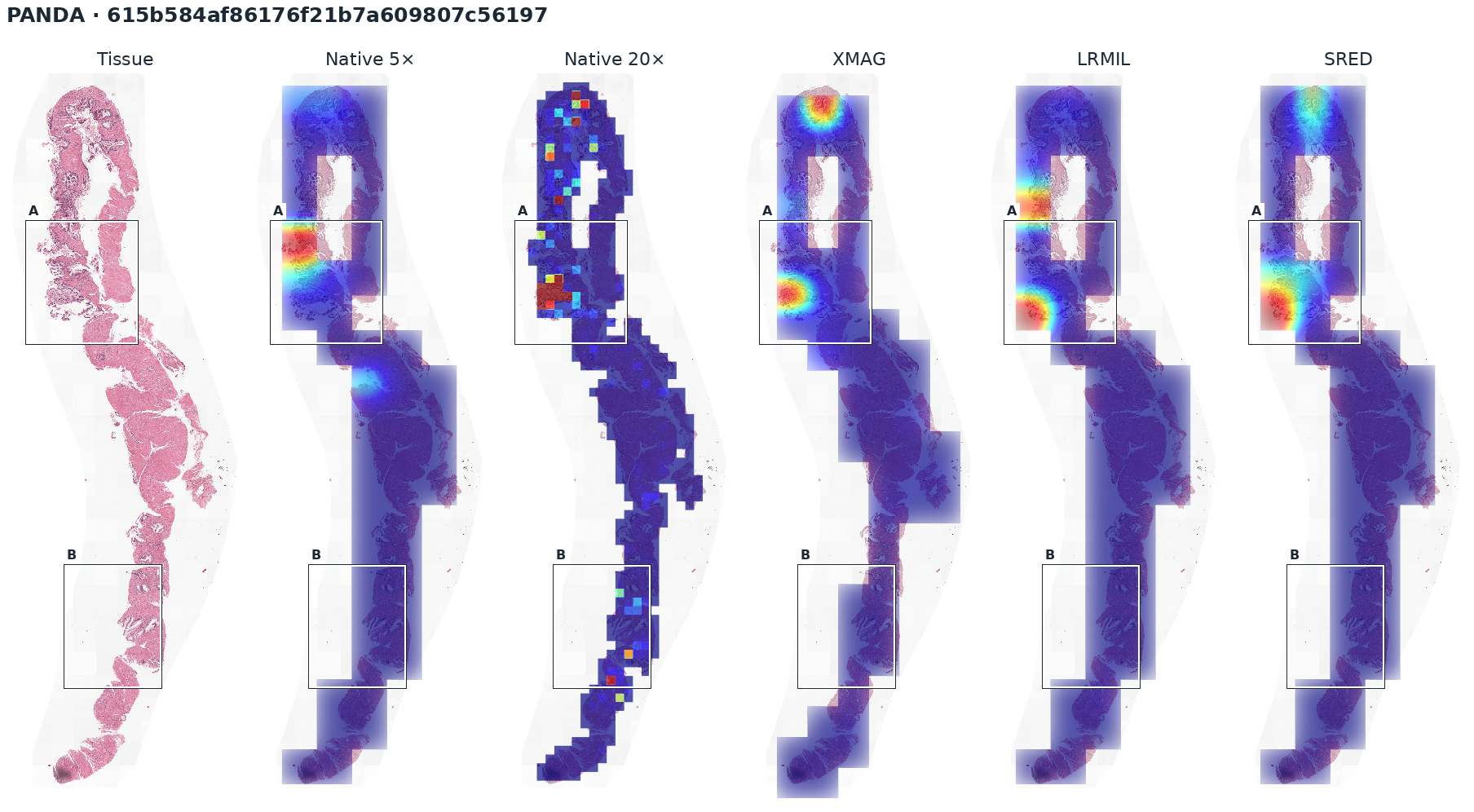}\\[3pt]
\includegraphics[width=0.94\textwidth]{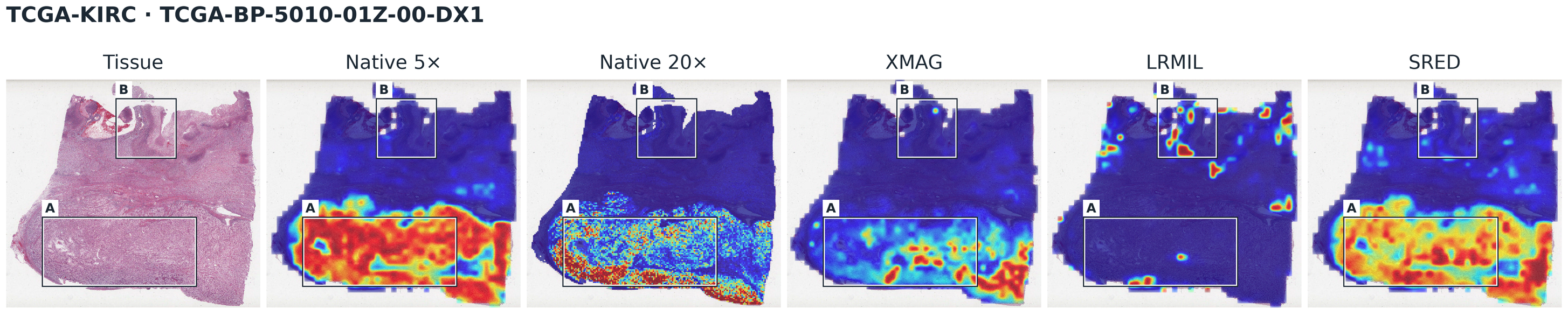}
\caption{\textbf{Attention on a prostate biopsy and a renal section.} Top: PANDA slide 615b584af86176f21b7a609807c56197. Bottom: TCGA-KIRC slide TCGA-BP-5010. A and B mark corresponding tissue locations across all panels.}
\label{fig:supp_attention_panda_kirc}
\end{figure*}

The PANDA biopsy in Fig.~\ref{fig:supp_attention_panda_kirc} shows localized responses along a narrow tissue core. Native \five{}, LRMIL and \methodname{} have upper foci in A, with peaks at different positions. Native \twenty{} also highlights small locations farther down the core in B, where the \five{} maps show little response. On KIRC, both native encoders and \methodname{} emphasize the lower tissue band in A. XMAG focuses near the lower right, while LRMIL also emphasizes the upper boundary in B.

\FloatBarrier
\begin{figure*}[!b]
\centering
\begin{minipage}[t]{\columnwidth}
\centering
\includegraphics[width=\linewidth]{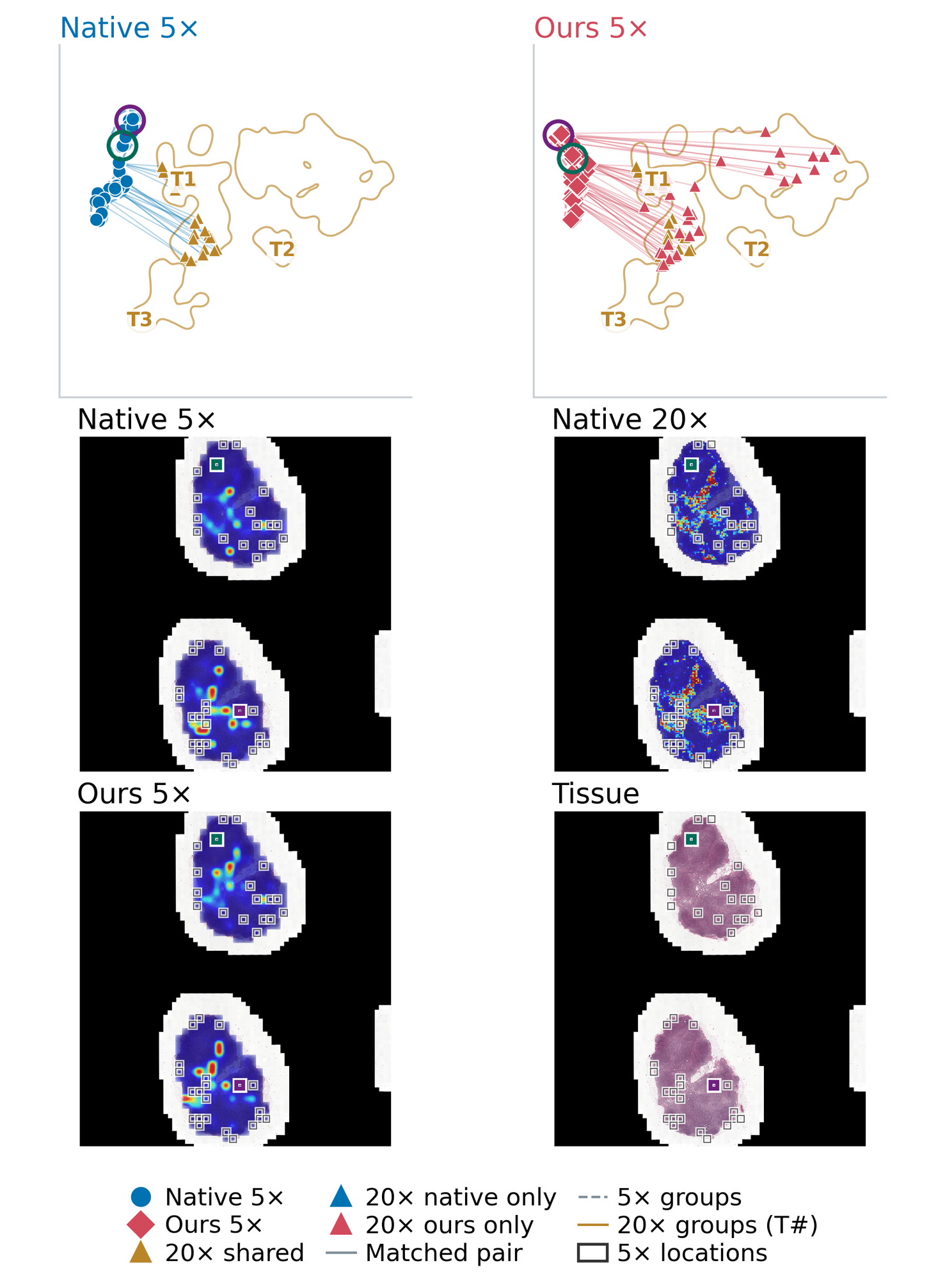}
\caption{\textbf{Feature correspondence and attention on CAM17.} Native (left) and distilled (right) features use the same selected \five{} patches. \methodname{} adds matches in T2. Colored outlines link corresponding parent patches across the feature plots and spatial maps.}
\label{fig:supp_correspondence_classification}
\end{minipage}\hfill
\begin{minipage}[t]{\columnwidth}
\centering
\includegraphics[width=\linewidth]{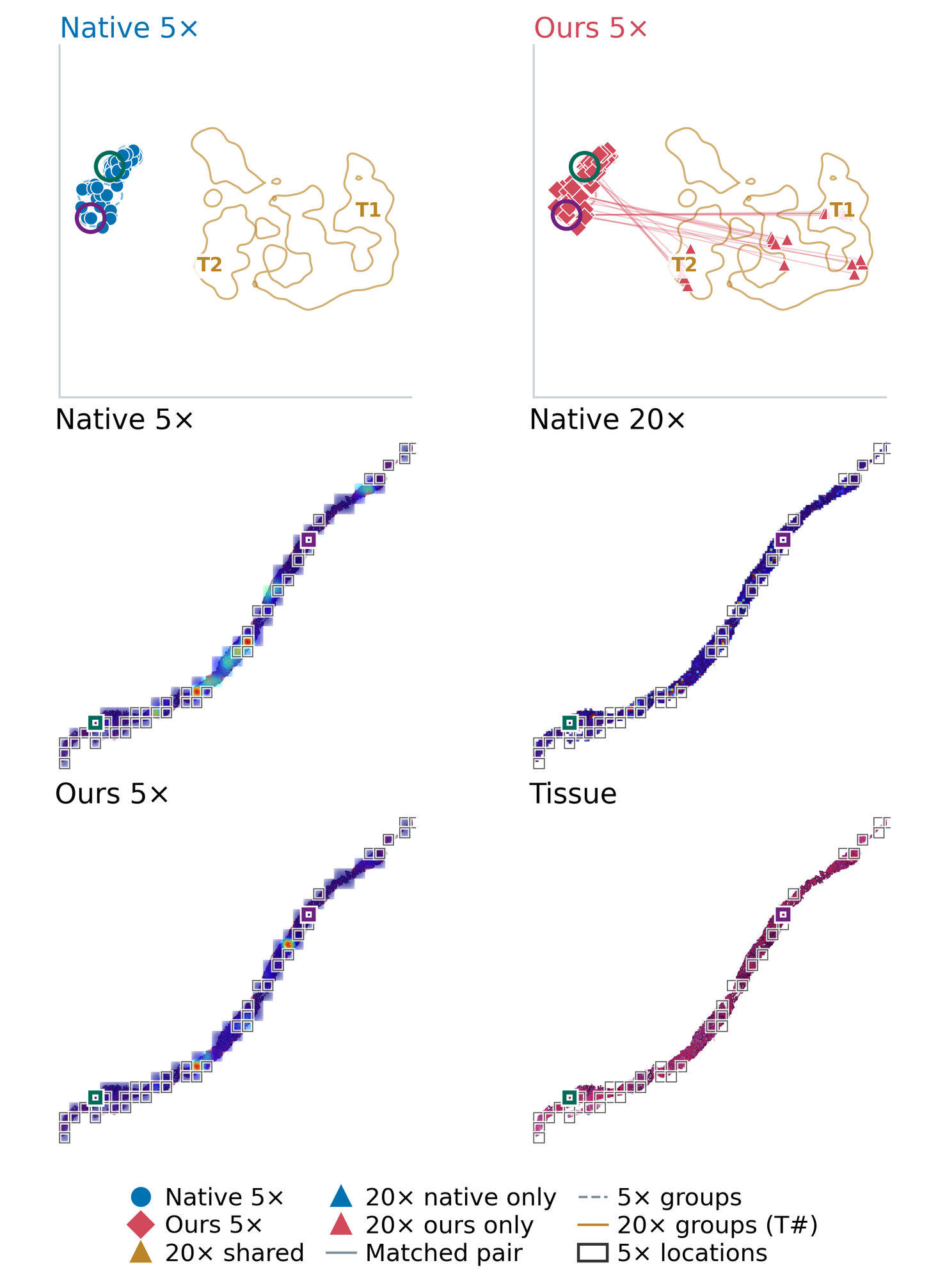}
\caption{\textbf{Feature correspondence and attention on PANDA.} Distillation adds matches in T1 and T2. The attention maps and tissue image show the spatial context of the selected features.}
\label{fig:supp_correspondence_panda}
\end{minipage}
\end{figure*}

\subsection{Feature correspondence and spatial responses}
\label{sec:supp_feature_correspondence}

The t-SNE plots~\cite{vandermaaten2008tsne} compare native and distilled features for the same \five{} patches, with lines connecting them to similar features from their aligned \twenty{} children. Paired tissue images and heatmaps show the selected patch locations and corresponding model responses. On CAM17 (Fig.~\ref{fig:supp_correspondence_classification}), \methodname{} adds matches in T2 and increases the number of matched children from 21 to 54. Both \five{} attention maps emphasize central areas of the two tissue sections, also prominent at \twenty{}. On PANDA (Fig.~\ref{fig:supp_correspondence_panda}), distilled features reach T1 and T2, with 21 matched children compared with none for native \five{}. The \methodname{} map preserves both prominent native \five{} responses. Native \twenty{} shows smaller foci along the biopsy.

\begin{figure*}[!tb]
\centering
\begin{minipage}[t]{\columnwidth}
\centering
\includegraphics[width=\linewidth]{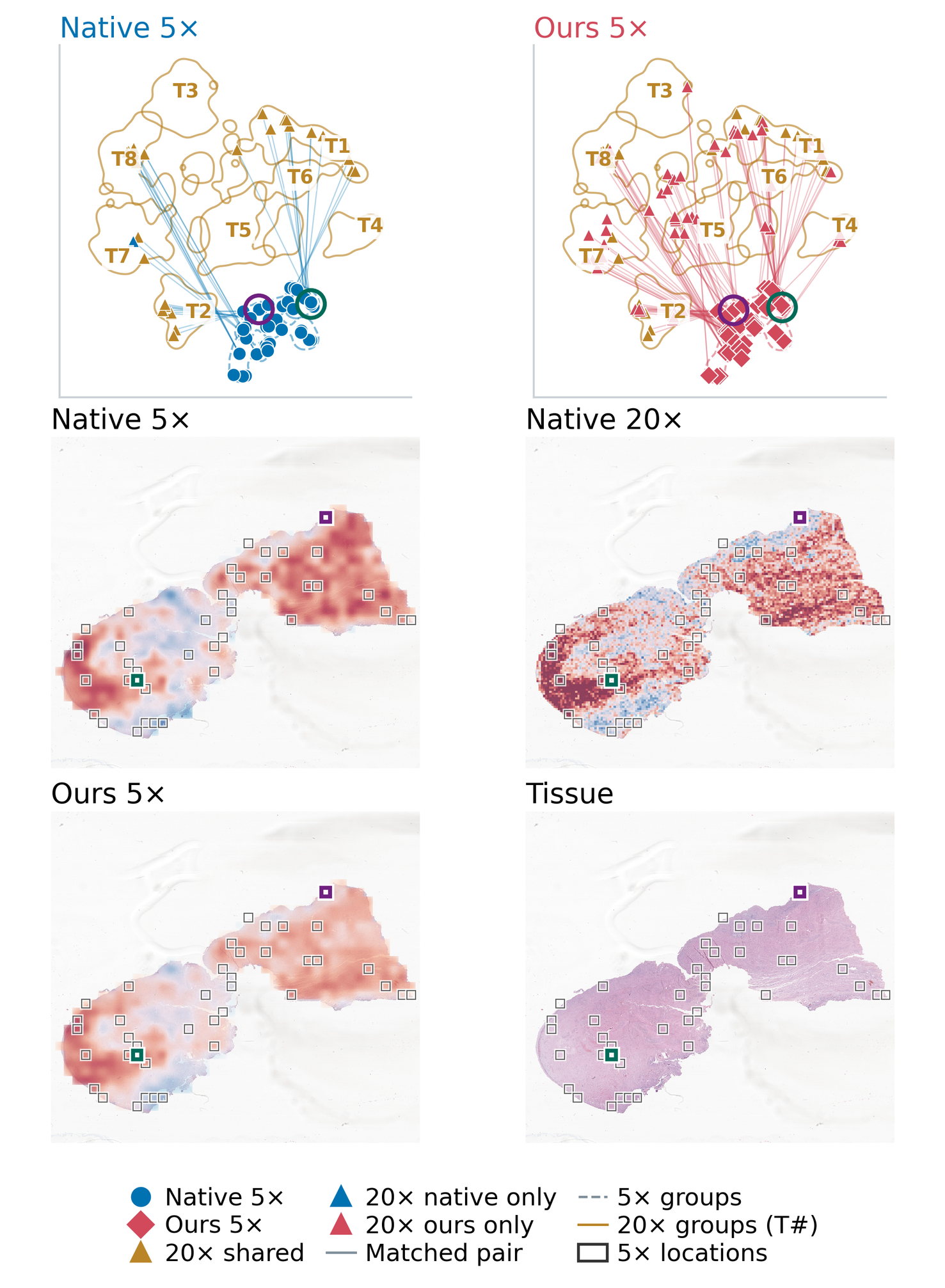}
\caption{\textbf{Feature correspondence and linear-probe responses on IMP-Cervix.} Distillation adds matches in T3, T4 and T5 on slide cervix\_221. Heatmaps show signed contributions to the annotated class logit. Colored outlines link the same tissue locations across panels.}
\label{fig:supp_correspondence_cervix_stad}
\end{minipage}\hfill
\begin{minipage}[t]{\columnwidth}
\centering
\includegraphics[width=\linewidth]{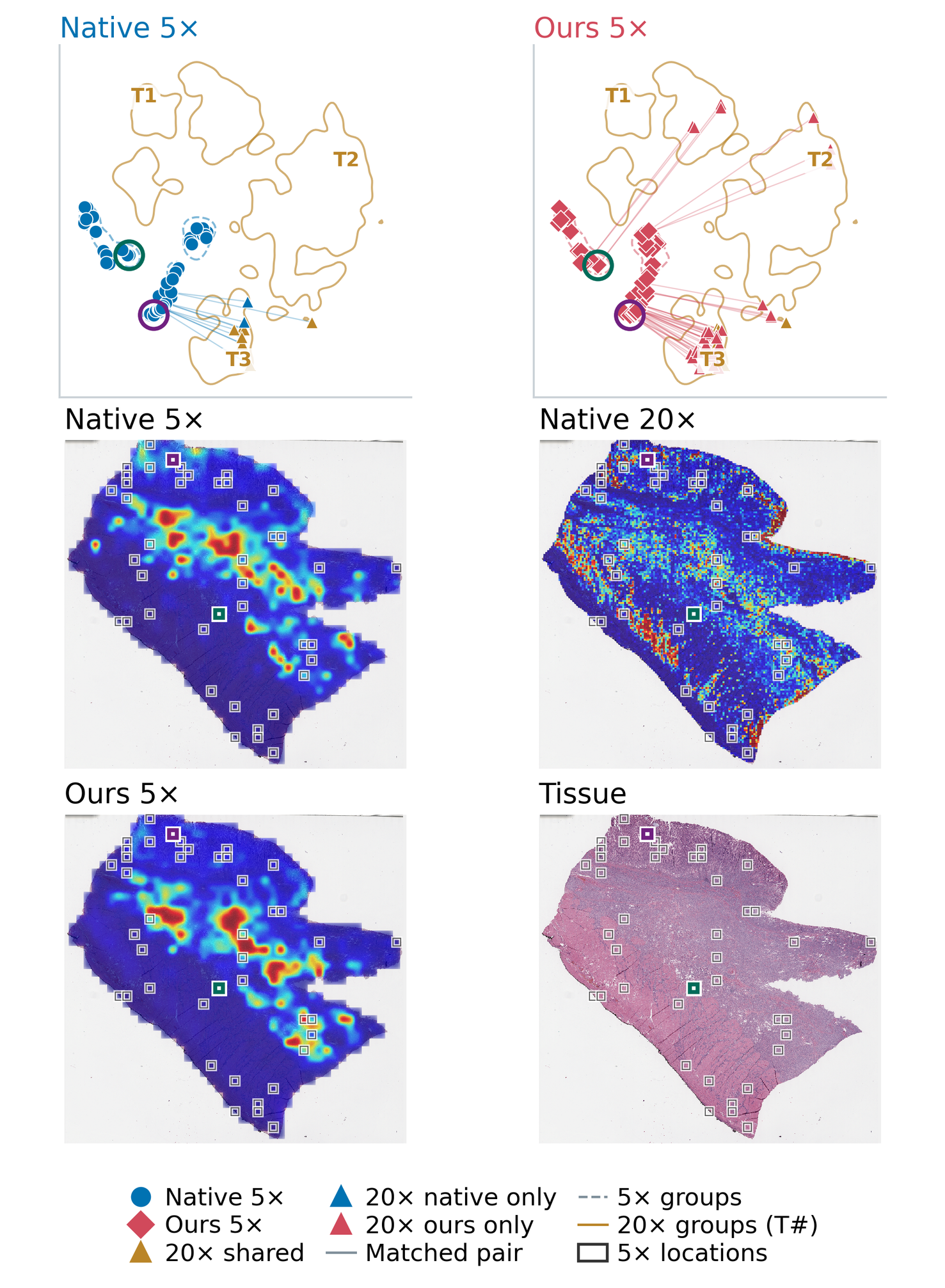}
\caption{\textbf{Feature correspondence and attention on TCGA-STAD.} On slide TCGA-VQ-A94U, distilled features reach all three teacher groups, compared with one for native \five{}. Colored outlines link corresponding parent patches across the feature plots and spatial maps.}
\label{fig:supp_correspondence_stad}
\end{minipage}
\end{figure*}

On IMP-Cervix (Fig.~\ref{fig:supp_correspondence_cervix_stad}), \methodname{} adds matches in T3, T4 and T5, increasing the total from 34 to 96. The maps show signed contributions to the annotated class logit from a linear probe. Red denotes positive contributions and blue negative contributions. Both \five{} maps show positive responses on the right and along the left fragment's boundary, with negative responses in its center and lower portion. On TCGA-STAD (Fig.~\ref{fig:supp_correspondence_stad}), \methodname{} adds matches in T1 and T2, reaching all three teacher groups, and increases the number of matched children from 10 to 38. Its attention map preserves the diagonal band seen at native \five{}. Native \twenty{} also emphasizes the left and lower edges.

%% file: main_arxiv.bbl
\begin{thebibliography}{61}
\providecommand{\natexlab}[1]{#1}
\providecommand{\url}[1]{\texttt{#1}}
\expandafter\ifx\csname urlstyle\endcsname\relax
  \providecommand{\doi}[1]{doi: #1}\else
  \providecommand{\doi}{doi: \begingroup \urlstyle{rm}\Url}\fi

\bibitem[Acosta et~al.(2022)Acosta, Panwar, Jarmale, Christie, Jasti, Margulis, Rakheja, Cheville, Leibovich, Parker, Brugarolas, Kapur, and Rajaram]{acosta2022heterogeneity}
Paul~H. Acosta, Vandana Panwar, Vipul Jarmale, Alana Christie, Jay Jasti, Vitaly Margulis, Dinesh Rakheja, John Cheville, Bradley~C. Leibovich, Alexander Parker, James Brugarolas, Payal Kapur, and Satwik Rajaram.
\newblock Intratumoral resolution of driver gene mutation heterogeneity in renal cancer using deep learning.
\newblock \emph{Cancer Research}, 82\penalty0 (15):\penalty0 2792--2806, 2022.

\bibitem[Alagha et~al.(2026)Alagha, Leclerc, Kotp, Metwally, Moras, Rentopoulos, Rostami, Nguyen, Baig, Khellaf, Trinh, Mizouni, Otrok, Bentahar, and Hosseini]{atlaspatch}
Ahmed Alagha, Christopher Leclerc, Yousef Kotp, Omar Metwally, Calvin Moras, Peter Rentopoulos, Ghodsiyeh Rostami, Bich~Ngoc Nguyen, Jumanah Baig, Abdelhakim Khellaf, Vincent Quoc-Huy Trinh, Rabeb Mizouni, Hadi Otrok, Jamal Bentahar, and Mahdi~S. Hosseini.
\newblock {AtlasPatch}: Scalable foundation model-based tissue detection and patch extraction for computational pathology.
\newblock \emph{arXiv preprint arXiv:2602.03998}, 2026.

\bibitem[Bergstrom et~al.(2024)Bergstrom, Abbasi, D{\'i}az-Gay, Galland, Ladoire, Lippman, and Alexandrov]{bergstrom2024hrd}
Erik~N. Bergstrom, Ammal Abbasi, Marcos D{\'i}az-Gay, Lo{\"i}ck Galland, Sylvain Ladoire, Scott~M. Lippman, and Ludmil~B. Alexandrov.
\newblock Deep learning artificial intelligence predicts homologous recombination deficiency and platinum response from histologic slides.
\newblock \emph{Journal of Clinical Oncology}, 42\penalty0 (30):\penalty0 3550--3560, 2024.

\bibitem[Boehm et~al.(2022)Boehm, Aherne, Ellenson, Nikolovski, Alghamdi, V{\'a}zquez-Garc{\'i}a, Zamarin, Long~Roche, Liu, Patel, Aukerman, Pasha, Rose, Selenica, Causa~Andrieu, Fong, Capanu, Reis-Filho, Vanguri, Veeraraghavan, Gangai, Sosa, Leung, McPherson, Gao, {MSK MIND Consortium}, Lakhman, and Shah]{boehm2022multimodal}
Kevin~M. Boehm, Emily~A. Aherne, Lora Ellenson, Ines Nikolovski, Mohammed Alghamdi, Ignacio V{\'a}zquez-Garc{\'i}a, Dmitriy Zamarin, Kara Long~Roche, Ying Liu, Druv Patel, Andrew Aukerman, Arfath Pasha, Doori Rose, Pier Selenica, Pamela~I. Causa~Andrieu, Chris Fong, Marinela Capanu, Jorge~S. Reis-Filho, Rami Vanguri, Harini Veeraraghavan, Natalie Gangai, Ramon Sosa, Samantha Leung, Andrew McPherson, JianJiong Gao, {MSK MIND Consortium}, Yulia Lakhman, and Sohrab~P. Shah.
\newblock Multimodal data integration using machine learning improves risk stratification of high-grade serous ovarian cancer.
\newblock \emph{Nature Cancer}, 3\penalty0 (6):\penalty0 723--733, 2022.

\bibitem[Brancati et~al.(2022)Brancati, Anniciello, Pati, Riccio, Scognamiglio, Jaume, De~Pietro, Di~Bonito, Foncubierta, Botti, Gabrani, Feroce, and Frucci]{brancati2022bracs}
Nadia Brancati, Anna~Maria Anniciello, Pushpak Pati, Daniel Riccio, Giosu{\`e} Scognamiglio, Guillaume Jaume, Giuseppe De~Pietro, Maurizio Di~Bonito, Antonio Foncubierta, Gerardo Botti, Maria Gabrani, Florinda Feroce, and Maria Frucci.
\newblock {BRACS}: A dataset for {BReAst} carcinoma subtyping in h\&e histology images.
\newblock \emph{Database}, 2022:\penalty0 baac093, 2022.

\bibitem[Bulten et~al.(2022)Bulten, Kartasalo, Chen, Str{\"o}m, Pinckaers, Nagpal, Cai, Steiner, van Boven, Vink, Hulsbergen-van~de Kaa, van~der Laak, Amin, Evans, van~der Kwast, Allan, Humphrey, Gr{\"o}nberg, Samaratunga, Delahunt, Tsuzuki, H{\"a}kkinen, Egevad, Demkin, Dane, Tan, Valkonen, Corrado, Peng, Mermel, Ruusuvuori, Litjens, Eklund, {the PANDA challenge consortium}, Brilhante, {\c{C}}ak{\i}r, Farr{\'e}, Geronatsiou, Molini{\'e}, Pereira, Roy, Saile, Salles, Schaafsma, Tschui, Billoch-Lima, Pereira, Zhou, He, Song, Sun, Yoshihara, Yamaguchi, Ono, Shen, Ji, Roussel, Zhou, Chai, Weng, Grechka, Shugaev, Kiminya, Kovalev, Voynov, Malyshev, Lapo, Campos, Ota, Yamaoka, Fujimoto, Yoshioka, Juvonen, Tukiainen, Karlsson, Guo, Hsieh, Zubarev, Bukhar, Li, Li, Speier, Arnold, Kim, Bae, Kim, Lee, and Park]{bulten2022panda}
Wouter Bulten, Kimmo Kartasalo, Po-Hsuan~Cameron Chen, Peter Str{\"o}m, Hans Pinckaers, Kunal Nagpal, Yuannan Cai, David~F. Steiner, Hester van Boven, Robert Vink, Christina Hulsbergen-van~de Kaa, Jeroen van~der Laak, Mahul~B. Amin, Andrew~J. Evans, Theodorus van~der Kwast, Robert Allan, Peter~A. Humphrey, Henrik Gr{\"o}nberg, Hemamali Samaratunga, Brett Delahunt, Toyonori Tsuzuki, Tomi H{\"a}kkinen, Lars Egevad, Maggie Demkin, Sohier Dane, Fraser Tan, Masi Valkonen, Greg~S. Corrado, Lily Peng, Craig~H. Mermel, Pekka Ruusuvuori, Geert Litjens, Martin Eklund, {the PANDA challenge consortium}, Am{\'e}rico Brilhante, Asl{\i} {\c{C}}ak{\i}r, Xavier Farr{\'e}, Katerina Geronatsiou, Vincent Molini{\'e}, Guilherme Pereira, Paromita Roy, G{\"u}nter Saile, Paulo G.~O. Salles, Ewout Schaafsma, Jo{\"e}lle Tschui, Jorge Billoch-Lima, Em{\'i}io~M. Pereira, Ming Zhou, Shujun He, Sejun Song, Qing Sun, Hiroshi Yoshihara, Taiki Yamaguchi, Kosaku Ono, Tao Shen, Jianyi Ji, Arnaud Roussel, Kairong Zhou, Tianrui Chai, Nina Weng, Dmitry Grechka, Maxim~V. Shugaev, Raphael Kiminya, Vassili Kovalev, Dmitry Voynov, Valery Malyshev, Elizabeth Lapo, Manuel Campos, Noriaki Ota, Shinsuke Yamaoka, Yusuke Fujimoto, Kentaro Yoshioka, Joni Juvonen, Mikko Tukiainen, Antti Karlsson, Rui Guo, Chia-Lun Hsieh, Igor Zubarev, Habib S.~T. Bukhar, Wenyuan Li, Jiayun Li, William Speier, Corey Arnold, Kyungdoc Kim, Byeonguk Bae, Yeong~Won Kim, Hong-Seok Lee, and Jeonghyuk Park.
\newblock Artificial intelligence for diagnosis and gleason grading of prostate cancer: the {PANDA} challenge.
\newblock \emph{Nature Medicine}, 28\penalty0 (1):\penalty0 154--163, 2022.

\bibitem[Chen et~al.(2022)Chen, Lu, Williamson, Chen, Lipkova, Noor, Shaban, Shady, Williams, Joo, and Mahmood]{chen2022pancancer}
Richard~J. Chen, Ming~Y. Lu, Drew F.~K. Williamson, Tiffany~Y. Chen, Jana Lipkova, Zahra Noor, Muhammad Shaban, Maha Shady, Mane Williams, Bumjin Joo, and Faisal Mahmood.
\newblock Pan-cancer integrative histology-genomic analysis via multimodal deep learning.
\newblock \emph{Cancer Cell}, 40\penalty0 (8):\penalty0 865--878.e6, 2022.

\bibitem[Chen et~al.(2024)Chen, Ding, Lu, Williamson, Jaume, Song, Chen, Zhang, Shao, Shaban, Williams, Oldenburg, Weishaupt, Wang, Vaidya, Le, Gerber, Sahai, Williams, and Mahmood]{chen2024uni}
Richard~J. Chen, Tong Ding, Ming~Y. Lu, Drew F.~K. Williamson, Guillaume Jaume, Andrew~H. Song, Bowen Chen, Andrew Zhang, Daniel Shao, Muhammad Shaban, Mane Williams, Lukas Oldenburg, Luca~L. Weishaupt, Judy~J. Wang, Anurag Vaidya, Long~Phi Le, Georg Gerber, Sharifa Sahai, Walt Williams, and Faisal Mahmood.
\newblock Towards a general-purpose foundation model for computational pathology.
\newblock \emph{Nature Medicine}, 30\penalty0 (3):\penalty0 850--862, 2024.

\bibitem[Chen et~al.(2020)Chen, Kornblith, Norouzi, and Hinton]{pmlr-v119-chen20j}
Ting Chen, Simon Kornblith, Mohammad Norouzi, and Geoffrey Hinton.
\newblock A simple framework for contrastive learning of visual representations.
\newblock In \emph{Proceedings of the 37th International Conference on Machine Learning}, pages 1597--1607. PMLR, 2020.

\bibitem[DiPalma et~al.(2021)DiPalma, Suriawinata, Tafe, Torresani, and Hassanpour]{dipalma2021resolution}
Joseph DiPalma, Arief~A. Suriawinata, Laura~J. Tafe, Lorenzo Torresani, and Saeed Hassanpour.
\newblock Resolution-based distillation for efficient histology image classification.
\newblock \emph{Artificial Intelligence in Medicine}, 119:\penalty0 102136, 2021.

\bibitem[Dong et~al.(2025)Dong, Jiang, Jiang, Li, and Zhang]{dong2025hdmil}
Jiuyang Dong, Junjun Jiang, Kui Jiang, Jiahan Li, and Yongbing Zhang.
\newblock Fast and accurate gigapixel pathological image classification with hierarchical distillation multi-instance learning.
\newblock In \emph{Proceedings of the IEEE/CVF Conference on Computer Vision and Pattern Recognition (CVPR)}, pages 30818--30828, 2025.

\bibitem[Ehteshami~Bejnordi et~al.(2017)Ehteshami~Bejnordi, Veta, van Diest, van Ginneken, Karssemeijer, Litjens, van~der Laak, et~al.]{bejnordi2017camelyon16}
Babak Ehteshami~Bejnordi, Mitko Veta, Paul~Johannes van Diest, Bram van Ginneken, Nico Karssemeijer, Geert Litjens, Jeroen A. W.~M. van~der Laak, et~al.
\newblock Diagnostic assessment of deep learning algorithms for detection of lymph node metastases in women with breast cancer.
\newblock \emph{JAMA}, 318\penalty0 (22):\penalty0 2199--2210, 2017.

\bibitem[Galland et~al.(2022)Galland, Ballot, Mananet, Boidot, Lecuelle, Albuisson, Arnould, Desmoulins, Mayeur, Kaderbhai, Ilie, Hennequin, Bergeron, Derang{\`e}re, Ghiringhelli, Truntzer, and Ladoire]{galland2022platinum}
Lo{\"i}ck Galland, Elise Ballot, Hugo Mananet, Romain Boidot, Julie Lecuelle, Juliette Albuisson, Laurent Arnould, Isabelle Desmoulins, Didier Mayeur, Cour{\`e}che Kaderbhai, Silvia Ilie, Audrey Hennequin, Anthony Bergeron, Valentin Derang{\`e}re, Fran{\c{c}}ois Ghiringhelli, Caroline Truntzer, and Sylvain Ladoire.
\newblock Efficacy of platinum-based chemotherapy in metastatic breast cancer and {HRD} biomarkers: utility of exome sequencing.
\newblock \emph{npj Breast Cancer}, 8\penalty0 (1):\penalty0 28, 2022.

\bibitem[Gao et~al.(2025)Gao, {Dupr\'e la Tour}, Tillman, Goh, Troll, Radford, Sutskever, Leike, and Wu]{gao2025scaling}
Leo Gao, Tom {Dupr\'e la Tour}, Henk Tillman, Gabriel Goh, Rajan Troll, Alec Radford, Ilya Sutskever, Jan Leike, and Jeffrey Wu.
\newblock Scaling and evaluating sparse autoencoders.
\newblock In \emph{International Conference on Learning Representations}, pages 26721--26754, 2025.

\bibitem[Han et~al.(2026)Han, Zhao, Deng, Huang, Liu, Lin, Lyu, Wang, Lu, Liang, Wen, Shi, Guo, and Liu]{han2026mag}
Chu Han, Bingchao Zhao, Tianpeng Deng, Jingqi Huang, Fangfang Liu, Jiatai Lin, Shanshan Lyu, Longfei Wang, Cheng Lu, Changhong Liang, Hannah~Y. Wen, Zhenwei Shi, Xiaojing Guo, and Zaiyi Liu.
\newblock A magnification alignment framework enables computation- and communication-efficient computational pathology.
\newblock \emph{Cancer Research}, 86\penalty0 (12):\penalty0 3060--3073, 2026.

\bibitem[Harrell et~al.(1982)Harrell, Califf, Pryor, Lee, and Rosati]{harrell1982evaluating}
Frank~E. Harrell, Jr., Robert~M. Califf, David~B. Pryor, Kerry~L. Lee, and Robert~A. Rosati.
\newblock Evaluating the yield of medical tests.
\newblock \emph{JAMA}, 247\penalty0 (18):\penalty0 2543--2546, 1982.

\bibitem[He et~al.(2026)He, Tessier-Cloutier, Wang, and Hosseini]{MF-MIL}
Haoyu He, Basile Tessier-Cloutier, Yang Wang, and Mahdi~S. Hosseini.
\newblock Masked feature encoding for large-scale whole slide image representation, 2026.

\bibitem[Heath et~al.(2021)Heath, Ferretti, Agrawal, An, Angelakos, Arya, Bajari, Baqar, Barnowski, Burt, Catton, Chan, Chu, Cullion, Davidsen, Do, Dompierre, Ferguson, Fitzsimons, Ford, Fukuma, Gaheen, Ganji, Garcia, George, Gerhard, Gerthoffert, Gomez, Han, Hernandez, Issac, Jackson, Jensen, Joshi, Kadam, Khurana, Kim, Kraft, Li, Lichtenberg, Lodato, Lolla, Martinov, Mazzone, Miller, Miller, Miller, Miyauchi, Murphy, Nullet, Ogwara, Ortu{\~{n}}o, Pedrosa, Pham, Popov, Porter, Powell, Rademacher, Reid, Rich, Rogel, Sahni, Savage, Schmitt, Simmons, Sislow, Spring, Stein, Sullivan, Tang, Thiagarajan, Troyer, Wang, Wang, West, Wilmer, Wilson, Wu, Wysocki, Xiang, Yamada, Yang, Yu, Yung, Zenklusen, Zhang, Zhang, Zhao, Zubair, Staudt, and Grossman]{heath2021gdc}
Allison~P. Heath, Vincent Ferretti, Stuti Agrawal, Maksim An, James~C. Angelakos, Renuka Arya, Rosita Bajari, Bilal Baqar, Justin H.~B. Barnowski, Jeffrey Burt, Ann Catton, Brandon~F. Chan, Fay Chu, Kim Cullion, Tanja Davidsen, Phuong-My Do, Christian Dompierre, Martin~L. Ferguson, Michael~S. Fitzsimons, Michael Ford, Miyuki Fukuma, Sharon Gaheen, Gajanan~L. Ganji, Tzintzuni~I. Garcia, Sameera~S. George, Daniela~S. Gerhard, Francois Gerthoffert, Fauzi Gomez, Kang Han, Kyle~M. Hernandez, Biju Issac, Richard Jackson, Mark~A. Jensen, Sid Joshi, Ajinkya Kadam, Aishmit Khurana, Kyle M.~J. Kim, Victoria~E. Kraft, Shenglai Li, Tara~M. Lichtenberg, Janice Lodato, Laxmi Lolla, Plamen Martinov, Jeffrey~A. Mazzone, Daniel~P. Miller, Ian Miller, Joshua~S. Miller, Koji Miyauchi, Mark~W. Murphy, Thomas Nullet, Rowland~O. Ogwara, Francisco~M. Ortu{\~{n}}o, Jes{\'u}s Pedrosa, Phuong~L. Pham, Maxim~Y. Popov, James~J. Porter, Raymond Powell, Karl Rademacher, Colin~P. Reid, Samantha Rich, Bessie Rogel, Himanso Sahni, Jeremiah~H. Savage, Kyle~A. Schmitt, Trevar~J. Simmons, Joseph Sislow, Jonathan Spring, Lincoln Stein, Sean Sullivan, Yajing Tang, Mathangi Thiagarajan, Heather~D. Troyer, Chang Wang, Zhining Wang, Bedford~L. West, Alex Wilmer, Shane Wilson, Kaman Wu, William~P. Wysocki, Linda Xiang, Joseph~T. Yamada, Liming Yang, Christine Yu, Christina~K. Yung, Jean~Claude Zenklusen, Junjun Zhang, Zhenyu Zhang, Yuanheng Zhao, Ariz Zubair, Louis~M. Staudt, and Robert~L. Grossman.
\newblock The nci genomic data commons.
\newblock \emph{Nature Genetics}, 53\penalty0 (3):\penalty0 257--262, 2021.

\bibitem[Hendrycks and Gimpel(2016)]{hendrycks2016gelu}
Dan Hendrycks and Kevin Gimpel.
\newblock Gaussian error linear units ({GELUs}).
\newblock \emph{arXiv preprint arXiv:1606.08415}, 2016.

\bibitem[Hinton et~al.(2015)Hinton, Vinyals, and Dean]{hinton2015distilling}
Geoffrey Hinton, Oriol Vinyals, and Jeff Dean.
\newblock Distilling the knowledge in a neural network.
\newblock \emph{arXiv preprint arXiv:1503.02531}, 2015.

\bibitem[Hosseini et~al.(2024)Hosseini, Bejnordi, Trinh, Chan, Hasan, Li, Yang, Kim, Zhang, Wu, Chinniah, Maghsoudlou, Zhang, Zhu, Khaki, Buin, Chaji, Salehi, Nguyen, Samaras, and Plataniotis]{hosseini2024computational}
Mahdi~S. Hosseini, Babak~Ehteshami Bejnordi, Vincent Quoc-Huy Trinh, Lyndon Chan, Danial Hasan, Xingwen Li, Stephen Yang, Taehyo Kim, Haochen Zhang, Theodore Wu, Kajanan Chinniah, Sina Maghsoudlou, Ryan Zhang, Jiadai Zhu, Samir Khaki, Andrei Buin, Fatemeh Chaji, Ala Salehi, Bich~Ngoc Nguyen, Dimitris Samaras, and Konstantinos~N. Plataniotis.
\newblock Computational pathology: A survey review and the way forward.
\newblock \emph{Journal of Pathology Informatics}, 15:\penalty0 100357, 2024.

\bibitem[Hou et~al.(2019)Hou, Pan, Loy, Wang, and Lin]{hou2019learning}
Saihui Hou, Xinyu Pan, Chen~Change Loy, Zilei Wang, and Dahua Lin.
\newblock Learning a unified classifier incrementally via rebalancing.
\newblock In \emph{Proceedings of the IEEE/CVF Conference on Computer Vision and Pattern Recognition (CVPR)}, pages 831--839, 2019.

\bibitem[Ilse et~al.(2018)Ilse, Tomczak, and Welling]{ilse2018attention}
Maximilian Ilse, Jakub Tomczak, and Max Welling.
\newblock Attention-based deep multiple instance learning.
\newblock In \emph{Proceedings of the 35th International Conference on Machine Learning}, pages 2127--2136. PMLR, 2018.

\bibitem[Javed et~al.(2023)Javed, Mahmood, Qaiser, and Werghi]{javed2023histology}
Sajid Javed, Arif Mahmood, Talha Qaiser, and Naoufel Werghi.
\newblock Knowledge distillation in histology landscape by multi-layer features supervision.
\newblock \emph{IEEE Journal of Biomedical and Health Informatics}, 27\penalty0 (4):\penalty0 2037--2046, 2023.

\bibitem[Kulkarni et~al.(2026)Kulkarni, Weng, Narayanaswamy, Liu, Sakla, and Thopalli]{kulkarni2026cbsae}
Akshay Kulkarni, Tsui-Wei Weng, Vivek Narayanaswamy, Shusen Liu, Wesam~A. Sakla, and Kowshik Thopalli.
\newblock Interpretable and steerable concept bottleneck sparse autoencoders.
\newblock In \emph{Proceedings of the IEEE/CVF Conference on Computer Vision and Pattern Recognition (CVPR)}, pages 2918--2927, 2026.

\bibitem[Le~Vuong and Kwak(2025)]{levuong2025moma}
Trinh~Thi Le~Vuong and Jin~Tae Kwak.
\newblock {MoMA}: Momentum contrastive learning with multi-head attention-based knowledge distillation for histopathology image analysis.
\newblock \emph{Medical Image Analysis}, 101:\penalty0 103421, 2025.

\bibitem[Li et~al.(2018)Li, Grandvalet, and Davoine]{pmlr-v80-li18a}
Xuhong Li, Yves Grandvalet, and Franck Davoine.
\newblock Explicit inductive bias for transfer learning with convolutional networks.
\newblock In \emph{Proceedings of the 35th International Conference on Machine Learning}, pages 2825--2834. PMLR, 2018.

\bibitem[Li et~al.(2023)Li, Dou, Da~Veiga~Leprevost, Geffen, Calinawan, Aguet, Akiyama, Anand, Birger, Cao, Chaudhary, Chilappagari, Cieslik, Colaprico, Zhou, Day, Domagalski, Esai~Selvan, Feny{\"o}, Foltz, Francis, Gonzalez-Robles, G{\"u}m{\"u}{\c{s}}, Heiman, Holck, Hong, Hu, Jaehnig, Ji, Jiang, Katsnelson, Ketchum, Klein, Lei, Liang, Liao, Lindgren, Ma, Ma, MacCoss, Martins~Rodrigues, McKerrow, Nguyen, Oldroyd, Pilozzi, Pugliese, Reva, Rudnick, Ruggles, Rykunov, Savage, Schnaubelt, Schraink, Shi, Singhal, Song, Storrs, Terekhanova, Thangudu, Thiagarajan, Wang, Wang, Wang, Wen, Wu, Wyczalkowski, Xin, Yao, Yi, Zhang, Zhang, Zuhl, Getz, Ding, Nesvizhskii, Wang, Robles, Zhang, Payne, and {Clinical Proteomic Tumor Analysis Consortium}]{li2023cptac}
Yize Li, Yongchao Dou, Felipe Da~Veiga~Leprevost, Yifat Geffen, Anna~P Calinawan, Fran{\c{c}}ois Aguet, Yo Akiyama, Shankara Anand, Chet Birger, Song Cao, Rekha Chaudhary, Padmini Chilappagari, Marcin Cieslik, Antonio Colaprico, Daniel~Cui Zhou, Corbin Day, Marcin~J Domagalski, Myvizhi Esai~Selvan, David Feny{\"o}, Steven~M Foltz, Alicia Francis, Tania Gonzalez-Robles, Zeynep~H G{\"u}m{\"u}{\c{s}}, David Heiman, Michael Holck, Runyu Hong, Yingwei Hu, Eric~J Jaehnig, Jiayi Ji, Wen Jiang, Lizabeth Katsnelson, Karen~A Ketchum, Robert~J Klein, Jonathan~T Lei, Wen-Wei Liang, Yuxing Liao, Caleb~M Lindgren, Weiping Ma, Lei Ma, Michael~J MacCoss, Fernanda Martins~Rodrigues, Wilson McKerrow, Ngoc Nguyen, Robert Oldroyd, Alexander Pilozzi, Pietro Pugliese, Boris Reva, Paul Rudnick, Kelly~V Ruggles, Dmitry Rykunov, Sara~R Savage, Michael Schnaubelt, Tobias Schraink, Zhiao Shi, Deepak Singhal, Xiaoyu Song, Erik Storrs, Nadezhda~V Terekhanova, Ratna~R Thangudu, Mathangi Thiagarajan, Liang-Bo Wang, Joshua~M Wang, Ying Wang, Bo Wen, Yige Wu, Matthew~A Wyczalkowski, Yi Xin, Lijun Yao, Xinpei Yi, Hui Zhang, Qing Zhang, Maya Zuhl, Gad Getz, Li Ding, Alexey~I Nesvizhskii, Pei Wang, Ana~I Robles, Bing Zhang, Samuel~H Payne, and {Clinical Proteomic Tumor Analysis Consortium}.
\newblock Proteogenomic data and resources for pan-cancer analysis.
\newblock \emph{Cancer Cell}, 41\penalty0 (8):\penalty0 1397--1406, 2023.

\bibitem[Litjens et~al.(2018)Litjens, Bandi, Ehteshami~Bejnordi, Geessink, Balkenhol, Bult, Halilovic, Hermsen, van~de Loo, Vogels, Manson, Stathonikos, Baidoshvili, van Diest, Wauters, van Dijk, and van~der Laak]{litjens2018camelyon}
Geert Litjens, Peter Bandi, Babak Ehteshami~Bejnordi, Oscar Geessink, Maschenka Balkenhol, Peter Bult, Altuna Halilovic, Meyke Hermsen, Rob van~de Loo, Rob Vogels, Quirine~F Manson, Nikolas Stathonikos, Alexi Baidoshvili, Paul van Diest, Carla Wauters, Marcory van Dijk, and Jeroen van~der Laak.
\newblock 1399 h\&e-stained sentinel lymph node sections of breast cancer patients: the {CAMELYON} dataset.
\newblock \emph{GigaScience}, 7\penalty0 (6):\penalty0 giy065, 2018.

\bibitem[Liu et~al.(2018)Liu, Lichtenberg, Hoadley, Poisson, Lazar, Cherniack, Kovatich, Benz, Levine, Lee, Omberg, Wolf, Shriver, Thorsson, {The Cancer Genome Atlas Research Network}, and Hu]{liu2018clinical}
Jianfang Liu, Tara Lichtenberg, Katherine~A. Hoadley, Laila~M. Poisson, Alexander~J. Lazar, Andrew~D. Cherniack, Albert~J. Kovatich, Christopher~C. Benz, Douglas~A. Levine, Adrian~V. Lee, Larsson Omberg, Denise~M. Wolf, Craig~D. Shriver, Vesteinn Thorsson, {The Cancer Genome Atlas Research Network}, and Hai Hu.
\newblock An integrated {TCGA} pan-cancer clinical data resource to drive high-quality survival outcome analytics.
\newblock \emph{Cell}, 173\penalty0 (2):\penalty0 400--416.e11, 2018.

\bibitem[Luo et~al.(2026)Luo, Xiang, Ji, and Li]{luo2026nnmil}
Xiangde Luo, Jinxi Xiang, Yuanfeng Ji, and Ruijiang Li.
\newblock {nnMIL}: a generalizable multiple instance learning framework for computational pathology.
\newblock \emph{Nature Biomedical Engineering}, 2026.

\bibitem[Mallya et~al.(2026)Mallya, Mirabadi, Farahani, and Bashashati]{mallya2026histomilkd}
Mayur Mallya, Ali~Khajegili Mirabadi, Hossein Farahani, and Ali Bashashati.
\newblock {HistoMILKD}: A multiple instance learning based multi-teacher knowledge distillation framework for whole slide image classification.
\newblock In \emph{Proceedings of the IEEE/CVF Winter Conference on Applications of Computer Vision (WACV)}, pages 3390--3400, 2026.

\bibitem[{National Cancer Institute}(n.d.)]{tcga_resource}
{National Cancer Institute}.
\newblock {The Cancer Genome Atlas (TCGA)}.
\newblock \url{https://www.cancer.gov/tcga}, n.d.
\newblock Accessed August 15, 2026.

\bibitem[Neto et~al.(2022)Neto, Oliveira, Montezuma, Fraga, Monteiro, Ribeiro, Gon{\c{c}}alves, Pinto, and Cardoso]{neto2022imil4path}
Pedro~C. Neto, Sara~P. Oliveira, Diana Montezuma, Jo{\~a}o Fraga, Ana Monteiro, Liliana Ribeiro, Sofia Gon{\c{c}}alves, Isabel~M. Pinto, and Jaime~S. Cardoso.
\newblock {iMIL4PATH}: A semi-supervised interpretable approach for colorectal whole-slide images.
\newblock \emph{Cancers}, 14\penalty0 (10):\penalty0 2489, 2022.

\bibitem[Neto et~al.(2024{\natexlab{a}})Neto, Montezuma, Oliveira, Oliveira, Fraga, Monteiro, Monteiro, Ribeiro, Gon{\c{c}}alves, Reinhard, Zlobec, Pinto, and Cardoso]{neto2024interpretable}
Pedro~C. Neto, Diana Montezuma, Sara~P. Oliveira, Domingos Oliveira, Jo{\~a}o Fraga, Ana Monteiro, Jo{\~a}o Monteiro, Liliana Ribeiro, Sofia Gon{\c{c}}alves, Stefan Reinhard, Inti Zlobec, Isabel~M. Pinto, and Jaime~S. Cardoso.
\newblock An interpretable machine learning system for colorectal cancer diagnosis from pathology slides.
\newblock \emph{npj Precision Oncology}, 8\penalty0 (1):\penalty0 56, 2024{\natexlab{a}}.

\bibitem[Neto et~al.(2024{\natexlab{b}})Neto, Montezuma, Oliveira, Oliveira, Fraga, Monteiro, Ribeiro, Gon{\c{c}}alves, Reinhard, Zlobec, Pinto, and Cardoso]{neto2024impcrs}
Pedro~C. Neto, Diana Montezuma, Sara~P. Oliveira, Domingos Oliveira, Jo{\~a}o Fraga, Ana Monteiro, Liliana Ribeiro, Sofia Gon{\c{c}}alves, Stefan Reinhard, Inti Zlobec, Isabel~M. Pinto, and Jaime~S. Cardoso.
\newblock {IMP} whole-slide images of colorectal samples 2024.
\newblock INESC TEC dataset, 2024{\natexlab{b}}.

\bibitem[Oliveira et~al.(2021)Oliveira, Neto, Fraga, Montezuma, Monteiro, Monteiro, Ribeiro, Gon{\c{c}}alves, Pinto, and Cardoso]{oliveira2021cad}
Sara~P. Oliveira, Pedro~C. Neto, Jo{\~a}o Fraga, Diana Montezuma, Ana Monteiro, Jo{\~a}o Monteiro, Liliana Ribeiro, Sofia Gon{\c{c}}alves, Isabel~M. Pinto, and Jaime~S. Cardoso.
\newblock {CAD} systems for colorectal cancer from {WSI} are still not ready for clinical acceptance.
\newblock \emph{Scientific Reports}, 11\penalty0 (1):\penalty0 14358, 2021.

\bibitem[Oliveira et~al.(2023)Oliveira, Montezuma, Moreira, Oliveira, Neto, Monteiro, Monteiro, Ribeiro, Gon{\c{c}}alves, Pinto, and Cardoso]{oliveira2023cervical}
Sara~P. Oliveira, Diana Montezuma, Ana Moreira, Domingos Oliveira, Pedro~C. Neto, Ana Monteiro, Jo{\~a}o Monteiro, Liliana Ribeiro, Sofia Gon{\c{c}}alves, Isabel~M. Pinto, and Jaime~S. Cardoso.
\newblock A {CAD} system for automatic dysplasia grading on h{\&}e cervical whole-slide images.
\newblock \emph{Scientific Reports}, 13\penalty0 (1):\penalty0 3970, 2023.

\bibitem[Oliveira et~al.(2024)Oliveira, Montezuma, Moreira, Oliveira, Neto, Monteiro, Monteiro, Ribeiro, Gon{\c{c}}alves, Pinto, and Cardoso]{oliveira2024impcervix}
Sara~P. Oliveira, Diana Montezuma, Ana Moreira, Domingos Oliveira, Pedro~C. Neto, Ana Monteiro, Jo{\~a}o Monteiro, Liliana Ribeiro, Sofia Gon{\c{c}}alves, Isabel~M. Pinto, and Jaime~S. Cardoso.
\newblock {IMP} whole-slide images of cervical samples 2024.
\newblock INESC TEC dataset, 2024.

\bibitem[Oquab et~al.(2024)Oquab, Darcet, Moutakanni, Vo, Szafraniec, Khalidov, Fernandez, Haziza, Massa, El-Nouby, Assran, Ballas, Galuba, Howes, Huang, Li, Misra, Rabbat, Sharma, Synnaeve, Xu, J{\'e}gou, Mairal, Labatut, Joulin, and Bojanowski]{oquab2023dinov2}
Maxime Oquab, Timoth{\'e}e Darcet, Th{\'e}o Moutakanni, Huy~V. Vo, Marc Szafraniec, Vasil Khalidov, Pierre Fernandez, Daniel Haziza, Francisco Massa, Alaaeldin El-Nouby, Mahmoud Assran, Nicolas Ballas, Wojciech Galuba, Russell Howes, Po-Yao Huang, Shang-Wen Li, Ishan Misra, Michael Rabbat, Vasu Sharma, Gabriel Synnaeve, Hu Xu, Herv{\'e} J{\'e}gou, Julien Mairal, Patrick Labatut, Armand Joulin, and Piotr Bojanowski.
\newblock {DINOv2}: Learning robust visual features without supervision.
\newblock \emph{Transactions on Machine Learning Research}, 2024.

\bibitem[Park et~al.(2019)Park, Kim, Lu, and Cho]{Park_2019_CVPR}
Wonpyo Park, Dongju Kim, Yan Lu, and Minsu Cho.
\newblock Relational knowledge distillation.
\newblock In \emph{Proceedings of the IEEE/CVF Conference on Computer Vision and Pattern Recognition (CVPR)}, pages 3967--3976, 2019.

\bibitem[Rajamanoharan et~al.(2024)Rajamanoharan, Lieberum, Sonnerat, Conmy, Varma, Kram{\'a}r, and Nanda]{rajamanoharan2024jumprelu}
Senthooran Rajamanoharan, Tom Lieberum, Nicolas Sonnerat, Arthur Conmy, Vikrant Varma, J{\'a}nos Kram{\'a}r, and Neel Nanda.
\newblock Jumping ahead: Improving reconstruction fidelity with {JumpReLU} sparse autoencoders.
\newblock \emph{arXiv preprint arXiv:2407.14435}, 2024.

\bibitem[Ravi et~al.(2025)Ravi, Gabeur, Hu, Hu, Ryali, Ma, Khedr, R{\"a}dle, Rolland, Gustafson, Mintun, Pan, Alwala, Carion, Wu, Girshick, Doll{\'a}r, and Feichtenhofer]{ravi2025sam2}
Nikhila Ravi, Valentin Gabeur, Yuan-Ting Hu, Ronghang Hu, Chaitanya Ryali, Tengyu Ma, Haitham Khedr, Roman R{\"a}dle, Chloe Rolland, Laura Gustafson, Eric Mintun, Junting Pan, Kalyan~Vasudev Alwala, Nicolas Carion, Chao-Yuan Wu, Ross Girshick, Piotr Doll{\'a}r, and Christoph Feichtenhofer.
\newblock {SAM 2}: Segment anything in images and videos.
\newblock In \emph{International Conference on Learning Representations}, pages 28085--28128, 2025.

\bibitem[Ridnik et~al.(2021)Ridnik, Ben-Baruch, Zamir, Noy, Friedman, Protter, and Zelnik-Manor]{Ridnik_2021_ICCV}
Tal Ridnik, Emanuel Ben-Baruch, Nadav Zamir, Asaf Noy, Itamar Friedman, Matan Protter, and Lihi Zelnik-Manor.
\newblock Asymmetric loss for multi-label classification.
\newblock In \emph{Proceedings of the IEEE/CVF International Conference on Computer Vision (ICCV)}, pages 82--91, 2021.

\bibitem[Roetzer-Pejrimovsky et~al.(2022)Roetzer-Pejrimovsky, Moser, Atli, Vogel, Mercea, Prihoda, Gelpi, Haberler, H{\"o}ftberger, Hainfellner, Baumann, Langs, and Woehrer]{roetzer2022ebrains}
Thomas Roetzer-Pejrimovsky, Anna-Christina Moser, Baran Atli, Clemens~Christian Vogel, Petra~A. Mercea, Romana Prihoda, Ellen Gelpi, Christine Haberler, Romana H{\"o}ftberger, Johannes~A. Hainfellner, Bernhard Baumann, Georg Langs, and Adelheid Woehrer.
\newblock The digital brain tumour atlas, an open histopathology resource.
\newblock \emph{Scientific Data}, 9\penalty0 (1):\penalty0 55, 2022.

\bibitem[Romero et~al.(2015)Romero, Ballas, Kahou, Chassang, Gatta, and Bengio]{romero2015fitnets}
Adriana Romero, Nicolas Ballas, Samira~Ebrahimi Kahou, Antoine Chassang, Carlo Gatta, and Yoshua Bengio.
\newblock {FitNets}: Hints for thin deep nets.
\newblock In \emph{International Conference on Learning Representations}, 2015.

\bibitem[Sammut(2022)]{sammut2022therapydata}
Stephen~John Sammut.
\newblock Multi-omic machine learning predictor of breast cancer therapy response.
\newblock Zenodo dataset, 2022.

\bibitem[Sammut et~al.(2022)Sammut, Crispin-Ortuzar, Chin, Provenzano, Bardwell, Ma, Cope, Dariush, Dawson, Abraham, Dunn, Hiller, Thomas, Cameron, Bartlett, Hayward, Pharoah, Markowetz, Rueda, Earl, and Caldas]{sammut2022therapy}
Stephen-John Sammut, Mireia Crispin-Ortuzar, Suet-Feung Chin, Elena Provenzano, Helen~A. Bardwell, Wenxin Ma, Wei Cope, Ali Dariush, Sarah-Jane Dawson, Jean~E. Abraham, Janet Dunn, Louise Hiller, Jeremy Thomas, David~A. Cameron, John M.~S. Bartlett, Larry Hayward, Paul~D. Pharoah, Florian Markowetz, Oscar~M. Rueda, Helena~M. Earl, and Carlos Caldas.
\newblock Multi-omic machine learning predictor of breast cancer therapy response.
\newblock \emph{Nature}, 601\penalty0 (7894):\penalty0 623--629, 2022.

\bibitem[Shao et~al.(2021)Shao, Bian, Chen, Wang, Zhang, Ji, and Zhang]{shao2021transmil}
Zhuchen Shao, Hao Bian, Yang Chen, Yifeng Wang, Jian Zhang, Xiangyang Ji, and Yongbing Zhang.
\newblock {TransMIL}: Transformer based correlated multiple instance learning for whole slide image classification.
\newblock In \emph{Advances in Neural Information Processing Systems}, pages 2136--2147. Curran Associates, Inc., 2021.

\bibitem[Shin and Jeong(2027)]{shin2026lrmil}
Yonghan Shin and Won-Ki Jeong.
\newblock {LRMIL}: Efficient low-resolution multiple instance learning via high-resolution knowledge distillation for whole slide image classification.
\newblock In \emph{Medical Image Computing and Computer Assisted Intervention -- MICCAI 2026}, pages 289--298. Springer Nature Switzerland, 2027.

\bibitem[Stanton et~al.(2021)Stanton, Izmailov, Kirichenko, Alemi, and Wilson]{doesKDwork2021}
Samuel Stanton, Pavel Izmailov, Polina Kirichenko, Alexander~A Alemi, and Andrew~G Wilson.
\newblock Does knowledge distillation really work?
\newblock In \emph{Advances in Neural Information Processing Systems}, pages 6906--6919. Curran Associates, Inc., 2021.

\bibitem[Su et~al.(2027)Su, Akbar, Sajjad, Parwani, and Niazi]{su2026xmag}
Ziyu Su, Abdul~Rehman Akbar, Usama Sajjad, Anil~V. Parwani, and Muhammad Khalid~Khan Niazi.
\newblock Streamline pathology foundation model by cross-magnification distillation.
\newblock \emph{Medical Image Analysis}, 115:\penalty0 104352, 2027.

\bibitem[Sun et~al.(2026)Sun, van Midden, Litjens, and Baumgartner]{sun2025protomil}
Susu Sun, Dominique van Midden, Geert Litjens, and Christian~F. Baumgartner.
\newblock Prototype-based multiple instance learning for gigapixel whole slide image classification.
\newblock In \emph{Medical Image Computing and Computer Assisted Intervention -- MICCAI 2025}, pages 507--517. Springer Nature Switzerland, 2026.

\bibitem[Tang et~al.(2024)Tang, Zhou, Huang, Zhu, Zhang, and Liu]{tang2024rrt}
Wenhao Tang, Fengtao Zhou, Sheng Huang, Xiang Zhu, Yi Zhang, and Bo Liu.
\newblock Feature re-embedding: Towards foundation model-level performance in computational pathology.
\newblock In \emph{Proceedings of the IEEE/CVF Conference on Computer Vision and Pattern Recognition (CVPR)}, pages 11343--11352, 2024.

\bibitem[Tran et~al.(2021)Tran, Gabbouj, and Iosifidis]{tran2021srm}
Dat~Thanh Tran, Moncef Gabbouj, and Alexandros Iosifidis.
\newblock Knowledge distillation by sparse representation matching.
\newblock \emph{arXiv preprint arXiv:2103.17012}, 2021.

\bibitem[van~der Maaten and Hinton(2008)]{vandermaaten2008tsne}
Laurens van~der Maaten and Geoffrey Hinton.
\newblock Visualizing data using {t-SNE}.
\newblock \emph{Journal of Machine Learning Research}, 9\penalty0 (86):\penalty0 2579--2605, 2008.

\bibitem[Wang et~al.(2021)Wang, Chang, Lo, Lin, Liou, Hsu, Lee, and Chao]{wang2021ovariandata}
Ching-Wei Wang, Chang-Cheng Chang, Shih-Chang Lo, Yi-Jia Lin, Yi-An Liou, Po-Chao Hsu, Yu-Ching Lee, and Tai-Kuang Chao.
\newblock A dataset of histopathological whole slide images for classification of treatment effectiveness to ovarian cancer ({Ovarian Bevacizumab Response}).
\newblock The Cancer Imaging Archive dataset, 2021.

\bibitem[Wang et~al.(2022)Wang, Chang, Khalil, Lin, Liou, Hsu, Lee, Wang, and Chao]{wang2022ovarian}
Ching-Wei Wang, Cheng-Chang Chang, Muhammad~Adil Khalil, Yi-Jia Lin, Yi-An Liou, Po-Chao Hsu, Yu-Ching Lee, Chih-Hung Wang, and Tai-Kuang Chao.
\newblock Histopathological whole slide image dataset for classification of treatment effectiveness to ovarian cancer.
\newblock \emph{Scientific Data}, 9\penalty0 (1):\penalty0 25, 2022.

\bibitem[Wilkinson et~al.(2020)Wilkinson, Ye, Karzai, Harmon, Terrigino, VanderWelle, Bright, Atway, Trostel, Carrabba, Whitlock, Walker, Lis, Sater, Capaldo, Madan, Gulley, Chun, Merino, Pinto, Salles, Kaur, Lotan, Venzon, Choyke, Turkbey, Dahut, and Sowalsky]{wilkinson2020nadt}
Scott Wilkinson, Huihui Ye, Fatima Karzai, Stephanie~A. Harmon, Nicholas~T. Terrigino, David~J. VanderWelle, John~R. Bright, Rayann Atway, Shana~Y. Trostel, Nicole~V. Carrabba, Nichelle~C. Whitlock, Stephanie~M. Walker, Rosina~T. Lis, Houssein~A. Sater, Brian~J. Capaldo, Ravi~A. Madan, James~L. Gulley, Guinevere Chun, Maria~J. Merino, Peter~A. Pinto, Daniela~C. Salles, Harsimar~B. Kaur, Tamara~L. Lotan, David~J. Venzon, Peter~L. Choyke, Baris Turkbey, William~L. Dahut, and Adam~G. Sowalsky.
\newblock Nascent prostate cancer heterogeneity drives evolution and resistance to intense hormonal therapy.
\newblock The Cancer Imaging Archive dataset, 2020.

\bibitem[Yang et~al.(2026)Yang, Cheng, and Hosseini]{yang2026transferable}
Zhiyuan Yang, Jiahao Cheng, and Mahdi~S. Hosseini.
\newblock What makes high-magnification knowledge transferable? a study of cross-resolution distillation in whole-slide imaging.
\newblock \emph{arXiv preprint arXiv:2609.36407}, 2026.

\bibitem[Zhang et~al.(2026)Zhang, Hao, Chen, Xu, Cong, Lu, and Xu]{zhang2025multimodal}
Qibin Zhang, Xinyu Hao, Qiao Chen, Rui Xu, Fengyu Cong, Cheng Lu, and Hongming Xu.
\newblock Multi-modal knowledge decomposition based online distillation for biomarker prediction in breast cancer histopathology.
\newblock In \emph{Medical Image Computing and Computer Assisted Intervention -- MICCAI 2025}, pages 353--363. Springer Nature Switzerland, 2026.

\end{thebibliography}
